\documentclass[11pt]{article}

\usepackage{n2s_arxiv}
\usepackage{xcolor}
\usepackage{tcolorbox}

\newcommand{\systemname}{\textsc{Notes2Skills}}
\newcommand{\metaskill}{\textsc{MetaSkill}}
\providecommand{\textcelsius}{\ensuremath{^\circ\mathrm{C}}}
\newcommand{\field}[1]{\texttt{#1}}
\newcommand{\fact}{\texttt{FACT}}
\newcommand{\judgment}{\texttt{JUDGMENT}}
\newcommand{\suggestion}{\texttt{SUGGESTION}}

\title{Notes2Skills: From Lab Notebooks to Certainty-Aware Scientific Agent Skills}
\author{Shi Liu \and Jiayao Chen \and Chengwei Qin \and Yanqing Hu \and Jufan Zhang \and Linyi Yang}
\date{September 2026}

\newcommand{\reporttitle}{\systemname: From Lab Notebooks to Certainty-Aware\\ Scientific Agent Skills}
\newcommand{\reportauthors}{Shi Liu$^{1,4}$ \quad Jiayao Chen$^1$ \quad Chengwei Qin$^2$ \quad Yanqing Hu$^1$\\[3pt]
Jufan Zhang$^3$ \quad Linyi Yang$^{1,\dagger}$}
\newcommand{\reportaffiliations}{
$^1$ Southern University of Science and Technology, Shenzhen, China\\
$^2$ The Hong Kong University of Science and Technology (Guangzhou), Guangzhou, China\\
$^3$ University College Dublin, Dublin, Ireland\\
$^4$ Shenzhen Loop Area Institute, Shenzhen, China}
\hypersetup{
  pdftitle={Notes2Skills: From Lab Notebooks to Certainty-Aware Scientific Agent Skills},
  pdfauthor={Shi Liu, Jiayao Chen, Chengwei Qin, Yanqing Hu, Jufan Zhang, Linyi Yang},
  pdfsubject={Extended technical report on certainty-preserving compilation of laboratory notes into scientific agent skills},
  pdfkeywords={Notes2Skills, lab notebooks, laboratory notes, tacit scientific knowledge, scientific agent skills, epistemic directive extraction, author certainty, skill compilation, source provenance, scientific data curation}}
\definecolor{n2sDeepRed}{HTML}{8C2F39}

\newcommand{\abskw}[1]{%
  \textbf{\textcolor{n2sDeepRed}{#1}}%
}

\newcommand{\reportversion}{%
  \begin{tabular}{@{}r@{}}
    {\itshape September 2026}\\[-1pt]
    {\scriptsize\bfseries\textcolor{n2sDeepRed}{Version 2}}\\[-2pt]
    {\scriptsize\textcolor{black!55}{Version 1: May 2026}}
  \end{tabular}%
}

\begin{document}

\paperheader{\reportversion}

\papertitle{\reporttitle}

\paperauthors{\reportauthors}

\paperaffil{\reportaffiliations}

\begin{abstract}
Scientific discovery workflows rely heavily on \abskw{lab notes}, where researchers record observations,
interpret uncertain results, and plan follow-up experiments. Unlike polished publications, lab notes preserve
\abskw{evolving scientific reasoning}, \abskw{tacit scientific knowledge}, and
\abskw{author uncertainty}, giving AI agents access to the process of science. However, most prior work on
scientific text focuses on papers, protocols, or structured databases, leaving informal laboratory notes
underexplored as inputs to AI agents for science. This gap matters because tacit knowledge in lab notes is not
written as a clean set of instructions: validated observations, tentative judgments, and possible experimental
next steps often appear in the same passage. If these signals are conflated, an AI agent may mistake uncertain
scientific judgments for confirmed conclusions or executable actions. To this end, we present \systemname{},
a two-stage framework for turning lab notebooks into verifiable skills for scientific AI agents while preserving
the author's certainty. Across three wet-lab sessions and seven original downstream configurations,
\systemname{} is the only tested configuration that avoids both observed regime-level failures:
over-acting on uncertain notes and missing firm directives. We show that
\abskw{certainty preservation} is a key missing piece between lab notebooks and reliable agent skills,
opening a path toward safer \abskw{AI co-scientist systems}.
\end{abstract}

\noindent
{\small
$\dagger$\,Correspondence:
\href{mailto:yangly6@sustech.edu.cn}
{\texttt{yangly6@sustech.edu.cn}}
}
\par

\vspace{5pt}

\noindent
{\footnotesize
\textcolor{black!58}{
This September 2026 technical report extends the May 2026 version of
\textsc{Notes2Skills}, which was accepted to the Findings of EMNLP 2026.
}}
\par

\vspace{7pt}

\section{Introduction}
\label{sec:introduction}

\begin{figure}[t]
  \centering
  \includegraphics[width=0.76\linewidth]{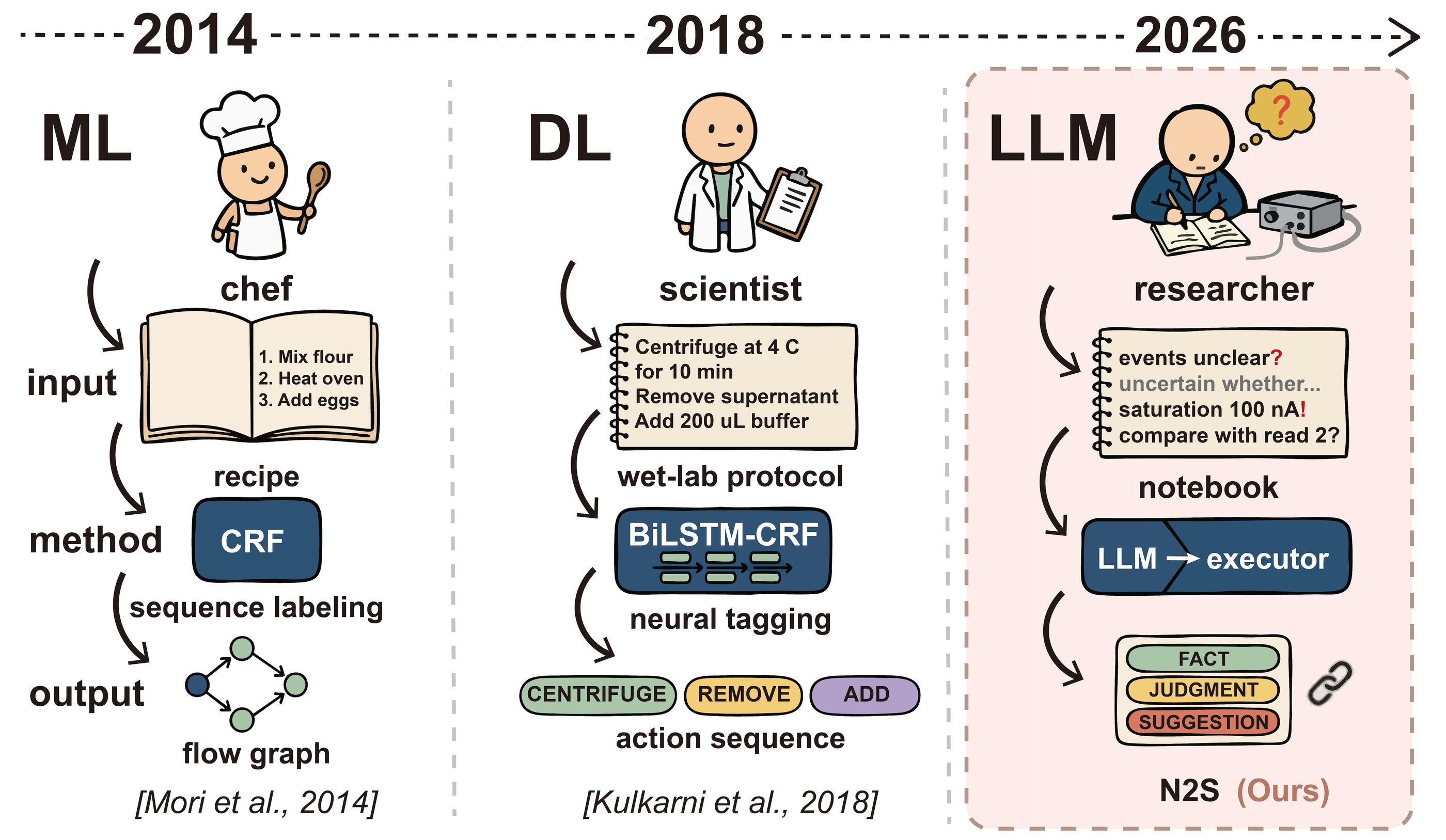}
  \caption{Three eras of procedural text extraction.}
  \label{fig:evolution}
\end{figure}

Lab notebooks are a central medium of scientific discovery, capturing researchers' observations, interpretations of uncertain results, and plans for subsequent experiments~\citep{doody2022labbook}. Unlike finalized papers and protocols, lab notebooks also preserve tacit knowledge in the scientist's judgments, qualifications, and choices during the experiment. Early work in the machine-learning era focused on extracting structured flow graphs from relatively constrained domains such as cooking recipes~\citep{mori_recipe_2014}. With the advent of neural sequence models, this line of research shifted toward deep-learning methods that identify and organize action sequences from published wet-lab protocols~\citep{wlp_2018,xwlp_2021,bioplanner_2023}. More recently, large language models (LLMs) have expanded procedural-text extraction and enabled workflows and domain knowledge to be packaged into agent-oriented workflows or skills~\citep{syntact_2025,anthropic_skills_2025}. In parallel, LLM-based scientific co-agents support hypothesis generation and experimental planning~\citep{gottweis_coscientist_2026}. Figure~\ref{fig:evolution} situates our notebook-to-skill setting in this progression from recipe-level flow graphs to protocol-level action modeling and LLM-based skill construction.

A common assumption underlies these three eras: the input text is prescriptive. A recipe instructs the reader to \emph{mix}; a wet-lab protocol specifies \emph{centrifuge for 10 minutes}; and a standard operating procedure states \emph{set the temperature to 4\textcelsius{}}. By the time such text is written, the author has typically resolved their uncertainty, leaving the extraction system to map explicit instructions into executable actions.

Experimental notebooks violate this assumption. They are written while the experiment is still unfolding, so
tacit knowledge often appears together with uncertainty rather
than as a settled instruction.

For example, \emph{the reading dropped sharply after five minutes} states a fact; \emph{I am not sure the second read is reliable} expresses a judgment under uncertainty; and \emph{try a fresh buffer next time} proposes a suggestion.
Although these statements should induce different downstream behaviors, their surface forms can appear deceptively similar in raw notebook text.

Treating them as if they were equally firm fails in two opposing failure modes.
\textbf{Uncertainty laundering} occurs when a tentative note, such as \emph{I am not sure the second read is reliable}, is compiled into a firm decision: the agent acts on an interpretation that the author explicitly marked as uncertain, potentially discarding underlying data based on an unresolved judgment. 
\textbf{Directive loss} captures the reverse failure: a firm directive, such as \emph{this part is invalid, truncate it}, is placed alongside cautious notes and treated as merely another opinion, causing the agent to retain data that the author intended to exclude.
Both failures arise when the compiler strips away the author's certainty signal: whether a statement is a fact, judgment, or suggestion. For agents performing irreversible operations on scientific data, this signal is the safety boundary.

In this work, we focus on single-author experimental notebooks written close to the time of experimentation, where the author's certainty directly informs decision-making. We study the tacit knowledge that becomes visible in how authors
state observations, qualify uncertain results, and propose next
steps. 
This regime poses unique challenges absent from published protocols: certainty is entangled with action content within the same sentence, and hedged judgments can be surface-similar to firm observations.

To address this, we present \systemname{} (N2S), motivated by the principle that the author's certainty should constrain what an agent may do: a \texttt{FACT} may license a strong processing action, a \texttt{JUDGMENT} defaults to conservative, 
review-preserving handling, and a \texttt{SUGGESTION} is treated as advisory only. In experiments, Stage~1 jointly extracts directives and their certainty labels; Stage~2 compiles them into a \metaskill{}, a Markdown skill document where each directive retains its certainty label and provenance to the original notebook excerpt.

We validate \systemname{} on 461 annotated segments across three corpora spanning a formality spectrum and test downstream skill loading on three real wet-lab sessions, where the compiled skills guide file-level data-handling decisions over instrument traces. On FreeNotes (informal bilingual notebooks), ONS (open notebook entries), and WLP (formal wet-lab protocols), the best of six model--prompt configurations achieves $F_1{=}0.737$ on binary directive detection, up from $0.682$ under the strongest zero-shot baseline, and our Stage~2 audit verifies that all 149 fixed directives are carried into source-linked, agent-loadable capsules. \systemname{} is the only tested configuration that avoids both observed failures: laundering uncertain readings into firm actions on uncertainty-heavy sessions, and losing firm author-stated actions on a \texttt{FACT}-dominated session.

To our knowledge, we are the first to consider transferring single-author experimental notebooks to agent-loadable skills. Our contributions are threefold as follows. First, we treat notebooks written by scientists as a new kind of procedural text, where \textbf{author certainty} serves as a safety boundary for agents. Second, we annotate 461 segments and audit 149 directives across three corpora with different levels of formality. Finally, we introduce \textbf{\metaskill{}} and show, on an aligned instrument-data benchmark, that certainty-preserving skills are needed to match expert-adjudicated triage grounded in the author's notes and signal evidence across both uncertain notes and firm facts.

\begin{figure*}[t]
  \centering

  \begin{minipage}[t]{0.32\textwidth}\vspace*{0pt}
    \begin{tcolorbox}[
      enhanced,
      equal height group=panels,
      colback=pagebg,
      colframe=pagebg,
      arc=2mm,
      boxrule=0pt,
      left=2.5mm,
      right=2.5mm,
      top=2mm,
      bottom=2mm
    ]
      {\small\itshape\bfseries Lab Notebook (FreeNotes)}
      \par\vspace{2pt}

      \begin{tcolorbox}[
        enhanced,
        colback=pagebg,
        frame hidden,
        borderline={0.7pt}{0pt}{dashgreen, dashed},
        arc=1.5mm,
        left=2mm,
        right=2mm,
        top=1.8mm,
        bottom=1.8mm,
        fontupper=\footnotesize\itshape
      ]
        \textbf{read1 @ 200\,mV:}
        events relatively sparse, baseline slightly drifting.
        \par\smallskip

        \textbf{Action:}
        replaced solution with PBS, cleared the pore.
        \par\smallskip

        \textbf{read2 @ 200\,mV:}
        events unclear,
        {\color{promptblue}uncertain whether a signal was detected}.
        \par\smallskip

        \textbf{Later:}
        current jumped to 100\,nA saturation;
        {\color{invalidred}signal invalid} from this point onward.
      \end{tcolorbox}
    \end{tcolorbox}
  \end{minipage}
  \hfill
  %
  \begin{minipage}[t]{0.32\textwidth}\vspace*{0pt}
    \begin{tcolorbox}[
      enhanced,
      equal height group=panels,
      colback=bgWLP,
      colframe=bgWLP,
      arc=2mm,
      boxrule=0pt,
      left=2.5mm,
      right=2.5mm,
      top=2mm,
      bottom=2mm
    ]
      {\small\itshape\bfseries Wet Lab Protocol (WLP)}
      \par\vspace{2pt}

      \begin{tcolorbox}[
        enhanced,
        colback=bgWLP,
        frame hidden,
        borderline={0.7pt}{0pt}{borderWLP, dashed},
        arc=1.5mm,
        left=2mm,
        right=2mm,
        top=1.8mm,
        bottom=1.8mm,
        fontupper=\footnotesize
      ]
        \textbf{Step 7.}
        Centrifuge at 12{,}000\,$\times$\,g for 10\,min at
        4\textcelsius{}.
        \par\smallskip

        \textbf{Note:}
        A visible pellet
        {\color{promptblue}should form}
        at the bottom of the tube.
        \par\smallskip

        {\color{invalidred}\textbf{DO NOT}}
        exceed 15\,min --- pellet may become too compact.
        \par\smallskip

        \textbf{OPTIONAL:}
        Supernatant may be kept at
        $-20$\textcelsius{} for up to 2 weeks.
      \end{tcolorbox}
    \end{tcolorbox}
  \end{minipage}
  \hfill
  %
  \begin{minipage}[t]{0.32\textwidth}\vspace*{0pt}
    \begin{tcolorbox}[
      enhanced,
      equal height group=panels,
      colback=bgONS,
      colframe=bgONS,
      arc=2mm,
      boxrule=0pt,
      left=2.5mm,
      right=2.5mm,
      top=2mm,
      bottom=2mm
    ]
      {\small\itshape\bfseries Open Notebook (ONS)}
      \par\vspace{2pt}

      \begin{tcolorbox}[
        enhanced,
        colback=bgONS,
        frame hidden,
        borderline={0.7pt}{0pt}{borderONS, dashed},
        arc=1.5mm,
        left=2mm,
        right=2mm,
        top=1.8mm,
        bottom=1.8mm,
        fontupper=\footnotesize
      ]
        Repeated yesterday's coupling reaction at the lower temperature.
        \par\smallskip

        TLC showed unreacted starting material ---
        {\color{promptblue}yield is likely too low to proceed}.
        \par\smallskip

        {\color{invalidred}This batch is unusable as-is};
        setting it aside.
        \par\smallskip

        \textit{Next time:}
        double the catalyst loading and extend the reaction to 24\,h.
      \end{tcolorbox}
    \end{tcolorbox}
  \end{minipage}

  \caption{
    Three scientific record genres express action-relevant information
    in distinct surface forms.
    Across informal lab notebooks, published wet-lab protocols, and
    open notebook entries, observations, uncertain judgments, firm
    assessments, procedural guidance, and prospective suggestions may
    appear side by side.
    Blue highlights hedged or qualified statements; red highlights
    firm action-relevant assessments or prohibitions.
  }
  \label{fig:notebook-genres}
\end{figure*}

\section{Lab Notebooks as a Source of Scientific Agent Skills}
\label{sec:notebook_skills}

\systemname{} studies a particular knowledge-transfer problem: how can an agent use a scientist's experimental notes without changing the strength of what the scientist actually said? The input is not an instruction manual awaiting execution. It is a record in which observations, interpretations, quality concerns, and prospective actions can coexist. A useful skill must preserve these distinctions, not merely retain the action words.

\subsection{Textual Traces of Tacit Scientific Knowledge}
\label{sec:tacit}

Tacit knowledge is part of scientific practice but is not fully captured by explicit accounts~\citep{polanyi1966tacit,collins2001tacit}. Laboratory notebooks provide a partial record of knowledge-making activity, including experiments, plans, procedures, and interpretations~\citep{doody2022labbook}. Our use of \emph{tacit scientific knowledge} concerns the part of that practice that becomes visible in notebook text: how an author qualifies a measurement, identifies a potentially unreliable region, or distinguishes an observation from a possible next step. We do not claim to recover expertise that the scientist never recorded.

This boundary matters for evaluation. A notebook may express a useful judgment without stating all of the experience that produced it. Preserving that judgment and its qualification does not reconstruct the author's complete reasoning. Conversely, a record need not be a polished instruction to affect downstream analysis. A statement that a reading is uncertain may justify retaining the data for review even though it does not authorize discarding them. The relevant object is therefore \emph{recorded guidance for later decisions}, not a complete inventory of a scientist's expertise.

The three examples in the Introduction illustrate this distinction. An observed drop, an uncertain assessment of a read, and a suggestion for the next experiment have different operational roles. These examples explain the task; they are not additional annotated records or a new evaluation set. The empirical measurements in this paper concern directive detection, directive type, epistemic status, compilation preservation, and file-level decisions. None is a direct measure of how much of a scientist's total tacit knowledge has been captured.

\subsection{Uncertainty Laundering and Directive Loss}
\label{sec:failure_definitions}

\paragraph{Uncertainty laundering.}
In notebook-to-skill processing, uncertainty laundering occurs when an author-qualified statement becomes a stronger downstream commitment without adequate authorization and corroborating evidence. The action need not be invented from nothing: a processing option may be present in the source, but its tentative status can disappear during extraction, compilation, or use. For example, uncertainty about whether a read is reliable does not by itself establish that the read should be excluded. The error is a change in what the source is allowed to authorize.

\paragraph{Directive loss.}
Directive loss is the opposing failure: a firm, applicable directive is omitted or weakened so that a supported processing action is missed. It can occur structurally, when a directive disappears from a skill, or operationally, when the directive remains visible but is treated as only an optional opinion. A blanket review policy may avoid some unsupported interventions while still failing to carry out firm instructions. Preserving a directive and executing it correctly are consequently different requirements.

These definitions motivate complementary checks. The compilation audit tests whether fixed input directives remain present with their labels and sources intact. The downstream experiment tests how the resulting representation and policy stack affect decisions. File-level review recall is informative about one vulnerable class, but it is not by itself a count of uncertainty-laundering events: a missed review can lead to different wrong actions. Likewise, high review recall does not establish that firm directives were followed. We therefore report the broader action-agreement metrics alongside review recall in Section~\ref{sec:results}.

\subsection{Author Certainty, Scientific Validity, and Action Commitment}
\label{sec:certainty_distinctions}

Three distinctions prevent the labels from acquiring stronger meanings than the task supports. \emph{Author certainty} records how the source author presents a statement. \emph{Scientific validity} concerns whether its content is correct. \emph{Action commitment} concerns the processing action authorized for a particular file under a domain policy and available evidence. A fourth quantity, a model's confidence in its prediction, concerns the extractor rather than the source author. It must not replace the author's epistemic label.

\begin{table}[htbp]
\centering\small
\begin{tabularx}{\linewidth}{@{}p{0.22\linewidth}YY@{}}
\toprule
\textbf{Concept} & \textbf{Question it answers} & \textbf{What it does not establish}\\
\midrule
Author certainty & How firmly did the author state this observation or directive? & That the statement is independently true.\\
Scientific validity & Is the observation, interpretation, or processing decision warranted? & That the source expressed it without qualification.\\
Action commitment & What handling is permitted for this file given the directive, policy, and evidence? & That the certainty label alone determines an action.\\
Model confidence & How confident is an extractor in a label or prediction? & How certain the original scientist was.\\
\bottomrule
\end{tabularx}
\caption{Distinct meanings of certainty and commitment in notebook-to-skill processing. The reported EDE task predicts author-side labels, not calibrated scientific truth or model confidence.}
\label{tab:certainty_distinctions}
\end{table}

In particular, \fact{} means an unhedged assertion by the author, not an independently verified scientific fact. \judgment{} records uncertainty, qualitative assessment, or tentative interpretation; its default is conservative handling. \suggestion{} records an optional or recommended action and is advisory. Their ordering is operational, not a universal psychological ranking. Additional file-level evidence may corroborate an already stored candidate without retroactively changing the author's original label. This distinction between preserving the source and checking a current action underlies the two-stage pipeline.

\section{Epistemic Directive Extraction from Laboratory Notes}
\label{sec:framework}\label{sec:stage1}

\begin{figure}[tbp]
\centering
\includegraphics[width=\linewidth]{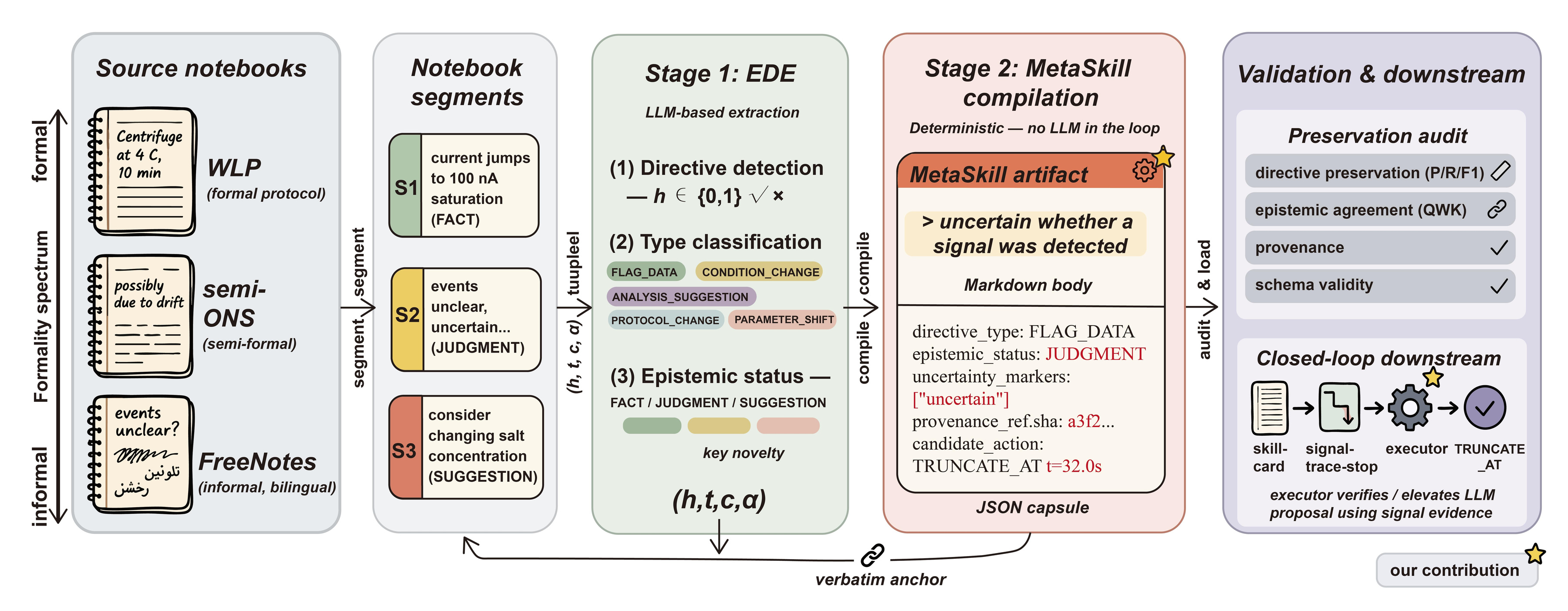}
\caption{Overview of \systemname{}. Stage~1 predicts directive presence, type, and epistemic status; Stage~2 deterministically compiles source-linked capsules. The right-hand panel separates preservation auditing from downstream execution. Example excerpts, candidate parameters, and abbreviated provenance strings are schematic, not additional evaluation records. A retained candidate is not an instruction to execute automatically.}
\label{fig:pipeline}
\end{figure}

\systemname{} separates predictive extraction from deterministic skill compilation. Stage~1, \emph{Epistemic Directive Extraction} (EDE), identifies statements that should guide later processing and labels their directive type and author certainty. Stage~2 carries these records into source-linked \metaskill{} capsules. Figure~\ref{fig:pipeline} summarizes this division; downstream evidence checking is a separate component rather than another extraction label.

\subsection{Prediction Target and Label Constraints}
\label{sec:ede_tuple}

For each input segment $s_i$, EDE predicts a tuple
\begin{equation}
 y_i=(h_i,t_i,c_i),\qquad h_i\in\{0,1\},
 \label{eq:ede}
\end{equation}
where $h_i$ denotes directive presence, $t_i$ is the directive type, and $c_i$ is the epistemic status. When $h_i=0$, both $t_i$ and $c_i$ must be null. When $h_i=1$, the type must belong to the five-category vocabulary in Table~\ref{tab:directive_types}, and $c_i\in\{\textsc{Fact},\textsc{Judgment},\textsc{Suggestion}\}$. The reported prompt asks for exactly these three fields as a single JSON object.

A directive is defined by its relevance to downstream pipeline behavior, not simply by grammatical mood. The shared instruction includes parameter changes, data-quality flags, analysis suggestions, protocol modifications, and condition changes. It excludes purely descriptive text, introductions, passing observations, and generic procedural steps without decision-changing content. The WLP corpus therefore serves as a protocol control under the same operational criterion; its inclusion does not imply that every imperative sentence is a positive example.

\begin{table}[htbp]
\centering
\small
\setlength{\tabcolsep}{5pt}
\renewcommand{\arraystretch}{1.12}
\begin{tabularx}{\linewidth}{@{}p{0.34\linewidth}Y@{}}
\toprule
\textbf{Type} & \textbf{Meaning} \\
\midrule
\texttt{FLAG\_DATA}           & Mark data that needs review, threshold change, or truncation. \\
\texttt{CONDITION\_CHANGE}    & Mark a phase or condition boundary. \\
\texttt{ANALYSIS\_SUGGESTION} & Suggest an analysis step. \\
\texttt{PROTOCOL\_CHANGE}     & Record a procedure change. \\
\texttt{PARAMETER\_SHIFT}     & Record a changed reference value. \\
\bottomrule
\end{tabularx}
\caption{The five directive types in Stage 1. Each type marks a different way a notebook statement can affect later analysis. \texttt{CONDITION\_CHANGE} is specific to FreeNotes; the other four span all three corpora.}
\label{tab:directive_types}
\end{table}

\subsection{Epistemic Status and Linguistic Context}
\label{sec:ede_context}

The label inventory distinguishes an unhedged author assertion from a tentative assessment and from optional advice. The shared prompt describes \judgment{} using uncertainty, qualitative assessment, or tentative interpretation, with examples such as ``seems'', ``may'', and ``looks like''. It describes \suggestion{} using optional or recommended action with user discretion. These cues help explain the annotation scheme, but a cue's presence is not a substitute for the statement's contextual meaning.

Corpus-specific framing complements the shared task. FreeNotes consists of informal Chinese--English notebook text, for which the prompt highlights bilingual hedge expressions. ONS is reflective and narrative, including introductions and summaries without actionable content. WLP is formal and predominantly imperative, with markers such as ``Note:'', ``Optional:'', and ``DO NOT''. Zero-shot and few-shot conditions share the same task definition. Few-shot messages additionally supply a fixed set of training-fold exemplars; their construction is documented in Section~\ref{sec:setup:stage1} and Appendix~\ref{app:prompts}.

These distinctions explain why directive type and certainty are evaluated separately. A data-quality flag can be tentative or firm. An analysis-related passage may describe what happened or recommend an optional next step. Predicting a plausible action category is therefore insufficient to recover the author's intended commitment. The observed errors at the \fact{}--\judgment{} boundary are examined in Section~\ref{sec:res:exp1}.

\subsection{From a Three-Label Prediction to a Source-Linked Record}
\label{sec:record_boundary}

The evaluated EDE prediction is narrower than the full record used for compilation. Source text and segment identity must remain available alongside the predicted labels. The capsule interface also includes uncertainty markers and directive scope, while the domain-policy layer can carry actions and typed parameters. These additional fields are inherited from the input directive record or supplied through domain configuration; they are not extra outputs of the three-label prompt in Equation~\ref{eq:ede}.

This separation is important when interpreting performance. Directive-detection or epistemic-status scores do not establish the accuracy of independently populated scopes, time references, or action parameters. Similarly, the 149-directive preservation audit uses fixed EDE records; it is not another estimate of the extractor's predictive accuracy. A faithful implementation must retain the distinction between the model's label output, the source-linked compilation input, and the domain-specific fields needed for execution. Section~\ref{sec:stage2} describes the corresponding paper-level interface without treating the three-label prediction as a complete executable record.

\section{\metaskill{}: Certainty-Preserving Skill Compilation}
\label{sec:stage2}

A \metaskill{} is an agent-loadable Markdown skill whose directives retain their source excerpts and epistemic labels. Its purpose is not to rewrite a notebook into uniformly confident instructions. It makes recorded guidance available while preserving the information needed to inspect and constrain its later use. The design follows four requirements: faithfulness to author certainty, machine-readable structure, source auditability, and conservative handling when action support is insufficient.

\subsection{Deterministic Compilation}
\label{sec:compile_mapping}

Let $\mathcal D=\{(s_i,t_i,c_i)\}_{i=1}^{|\mathcal D|}$ denote the positive EDE directives, where $s_i$ identifies the source segment. At this paper-level abstraction, additional source-record attributes accompany the tuple when needed by the domain adapter. Given compile-time domain configuration $\Pi$, Stage~2 applies
\begin{equation}
 M=\mathrm{Compile}(\mathcal D,\Pi).
 \label{eq:compile}
\end{equation}
The compiler is deterministic: each capsule field is inherited from the input record or fixed by configuration. There is no LLM in Stage~2. One capsule is emitted per fixed directive, with the original excerpt presented alongside its structured representation.

Determinism makes input-to-output preservation directly inspectable. It does not make the input labels correct, validate the scientific content of an excerpt, or establish that the configured policy is appropriate in every laboratory. An LLM-based compiler would need additional mechanisms to achieve the same exact preservation checks; our choice avoids making source fidelity depend on another generative step.

\subsection{Universal Content and Domain-Policy Fields}
\label{app:schema}\label{sec:capsule_schema}

Table~\ref{tab:metaskill_schema} gives the paper-level capsule interface. Universal fields preserve the directive's identity, meaning, certainty, and source. Domain-policy fields specify how that directive can be used in a particular action vocabulary. Build-only metadata is omitted from this interface. A source link enables comparison with the original segment; it is not a certificate of the statement's truth.

\begin{table}[htbp]
\centering\small
\setlength{\tabcolsep}{4pt}\renewcommand{\arraystretch}{1.14}
\begin{tabularx}{\linewidth}{@{}p{0.28\linewidth}YY@{}}
\toprule
\textbf{Field} & \textbf{Information retained or supplied} & \textbf{Role in inspection or use}\\
\midrule
\field{source\_segment\_id} & Identity of the original segment. & Locate the source excerpt.\\
\field{directive\_type} & EDE directive category. & Preserve the kind of downstream guidance.\\
\field{epistemic\_status} & Author-side EDE certainty label. & Keep the original commitment visible.\\
\field{uncertainty\_markers} & Verbatim uncertainty cues, when present in the input record. & Inspect how the source qualifies the statement.\\
\field{flag\_scope} & Scope recorded for a \field{FLAG\_DATA} directive. & Distinguish which part of the data the flag concerns.\\
\midrule
\multicolumn{3}{@{}l}{\textit{Domain-policy fields}}\\[2pt]
\field{default\_action} & Default licensed processing action. & Specify default handling, not every possible intervention.\\
\field{default\_action\_tier} & Commitment tier for authorization. & Support an action-specific gate.\\
\field{explicit\_authorization} & Structured support for a strong action. & Make the relevant authorization path inspectable.\\
\field{candidate\_actions} & Non-default actions retained for evidence matching. & Preserve options without making them automatic defaults.\\
\bottomrule
\end{tabularx}
\caption{Paper-level \metaskill{} interface, expanded from the camera-ready schema. Source-linked attributes accompany the three-label EDE prediction; policy fields come from the input record and domain configuration. This table describes their roles, not an additional evaluated field-extraction model.}
\label{tab:metaskill_schema}
\end{table}

The distinction between labels and supporting attributes also defines the scope of a provenance audit. The certainty label must agree with its fixed EDE input, the excerpt must remain faithful to the source, and domain fields must be interpretable under the configured action policy. Parameters already supplied by the source record or configuration retain their types. At runtime, the executor may check an existing parameter against the signal record, but it does not create a missing timestamp or threshold.

\subsection{Preservation Checks and Their Meaning}
\label{sec:preservation_contract}

The Stage~2 audit checks four universal properties: directive preservation, unchanged epistemic status, source fidelity, and schema validity. At the directive-identity level, the intended correspondence can be written as
\begin{equation}
 \operatorname{IDs}(M)=\operatorname{IDs}(\mathcal D),\qquad
 c_i^{M}=c_i^{\mathcal D}\quad\text{for every matched directive }i.
 \label{eq:preservation}
\end{equation}
Here $\operatorname{IDs}$ retains the multiplicity of each directive identity. With unique input directive keys, the first condition therefore requires one capsule per input directive rather than merely equal totals. The second preserves each label rather than only its corpus-level distribution. Source fidelity and schema checks then establish that the matching capsule points back to its excerpt and uses valid fields. FreeNotes additionally has an action-policy check; WLP and ONS do not use that downstream policy layer in the reported audit.

These checks answer a conditional question: given the fixed compilation input, did the skill preserve it? They do not answer whether Stage~1 found every directive in the original notebook. They also do not measure whether an agent subsequently obeyed the capsule. The extraction evaluation, compilation audit, and downstream decision experiment therefore have different units and cannot be substituted for one another. Their corresponding evidence is reported in Sections~\ref{sec:res:exp1}--\ref{sec:res:exp3}.

\subsection{Keeping a Candidate without Committing to It}
\label{sec:candidate_actions}

Preserving a directive does not require executing every action associated with it. \fact{} may support a strong action when policy and file-level evidence also authorize it. \judgment{} defaults to review-preserving handling while retaining a non-default candidate where the input record and policy provide one. \suggestion{} is advisory. For a context directive, the skill can use \field{KEEP\_FULL\_WITH\_NOTE}; the agent prompt maps this to the tool action \field{KEEP\_FULL} with the contextual content in the rationale, rather than introducing a sixth evaluated action.

A candidate therefore records an available option, not a newly validated conclusion. Its later corroboration concerns the current file; it does not replace \judgment{} with \fact{} in the source capsule. This permits source certainty to remain stable while an action is checked against additional evidence. The supported substitution rule is deliberately narrow and is specified in Section~\ref{sec:setup:executor}.

The combination of author-side commitment and signal-side checking admits a prior-like decision-theoretic interpretation: strong intervention is withheld unless the relevant authorization and evidence agree. This is only an interpretation of the rule. \systemname{} does not fit a Bayesian posterior, calibrate the three labels as probabilities, or assume statistical independence between the author and signal channels. The operative mechanism is the explicit domain-specific authorization check.

\section{Evidence-Grounded Execution for Scientific Data Curation}
\label{sec:execution}

The downstream task is file-level scientific data curation: deciding how to retain, review, or process an instrument recording. It is not an evaluation of autonomous wet-lab intervention or hypothesis discovery. In the reported nanopore workflow, a compiled skill contributes source-linked guidance, an agent proposes a processing decision, and a deterministic executor checks whether the required authorization and signal evidence support it.

\subsection{Signals, Proposals, and Checked Decisions}
\label{sec:setup:arch}

The pipeline has three layers. Layer~1 converts an instrument trace into a fixed \field{SignalFindings} summary. Layer~2 is a Claude Sonnet~4.5 agent loop that reads the supplied representation and per-file summary and proposes a processing decision through \field{emit\_decision}. Layer~3 checks the proposal against the capsule and signal evidence. This separation identifies which component produces a proposal and which component determines whether it passes the configured evidence checks.

The five actions are \field{KEEP\_FULL}, \field{FLAG\_FOR\_REVIEW}, \field{RAISE\_THRESHOLD}, \field{TRUNCATE\_AT}, and \field{SKIP\_FILE}. They respectively retain the complete recording, retain it for review, change event-detection thresholding, truncate at a specified time, or exclude the file from downstream processing. Parameterized actions require an existing structured parameter, such as \field{truncate\_at\_s} or \field{threshold\_multiplier}. Merely returning a valid action name is not enough to justify a parameterized intervention.

\subsection{The Domain Adapter}
\label{app:adapter}\label{sec:domain_adapter}

The reusable epistemic contract comprises EDE label semantics, source-linked compilation, and preservation checks. Execution additionally depends on a domain adapter
\begin{equation}
 \Pi_d=(A_d,\Theta_d,E_d,M_d,G_d,F_d),
 \label{eq:adapter}
\end{equation}
where $A_d$ is the action vocabulary, $\Theta_d$ the typed parameter schema, $E_d$ the evidence schema, $M_d$ the directive-to-action mapping, $G_d$ the authorization predicates, and $F_d$ the fail-closed fallback. The subscript distinguishes these adapter components from the compiled skill $M$ in Equation~\ref{eq:compile}. Only the nanopore adapter is instantiated and evaluated in this paper.

An action's support is specific to its scope and parameter. A general concern about signal quality cannot provide a missing truncation boundary. Nor does the existence of an intervention in one directive establish its applicability to every recording in a session. This is why the capsule, file-level evidence, and proposed action must be considered together. Source certainty limits default commitment, while the adapter determines what additional evidence is required for the current file.

\subsection{Authorization, Veto, and Candidate Elevation}
\label{sec:setup:executor}\label{app:check}

The executor distinguishes strong proposals---\field{RAISE\_THRESHOLD}, \field{TRUNCATE\_AT}, and \field{SKIP\_FILE}---from cautious proposals, \field{KEEP\_FULL} and \field{FLAG\_FOR\_REVIEW}. The \emph{verify} setting enables two operations. \textsc{Authorize} passes a strong proposal only when an action-specific capsule authorization path and the required signal evidence agree. \textsc{Veto} applies an action-specific fallback when that support is insufficient. Table~\ref{tab:executor_transitions} summarizes the latter; it is not a generic move to the next action on a severity scale.

\begin{table}[htbp]
\centering\small
\begin{tabularx}{\linewidth}{@{}p{0.24\linewidth}Yp{0.25\linewidth}@{}}
\toprule
\textbf{Unsupported proposal} & \textbf{Fallback condition} & \textbf{Output}\\
\midrule
\field{SKIP\_FILE} & Authorization or evidence is insufficient. & \field{FLAG\_FOR\_REVIEW}\\
\field{RAISE\_THRESHOLD} & Authorization, evidence, or parameter support is insufficient. & \field{FLAG\_FOR\_REVIEW}\\
\field{TRUNCATE\_AT} & An existing threshold intervention is independently supported. & \field{RAISE\_THRESHOLD}\\
\field{TRUNCATE\_AT} & Otherwise unsupported. & \field{FLAG\_FOR\_REVIEW}\\
\bottomrule
\end{tabularx}
\caption{Fail-closed transitions of the nanopore executor. The threshold fallback still requires its own existing, supported intervention; it does not create a new threshold parameter.}
\label{tab:executor_transitions}
\end{table}

The \emph{elevate} setting adds \textsc{Substitute} for an originally cautious Layer~2 proposal. It can substitute an existing, non-context \field{FLAG\_DATA} truncation candidate only when exactly one stored candidate timestamp matches a calibrated step drop. Conflicting candidates do not trigger substitution. If no unique candidate is corroborated, the cautious proposal is retained. This step does not search for arbitrary new interventions, synthesize a boundary, or reinterpret every tentative note as an actionable command.

The ordering of these checks matters. Candidate elevation is tied to the originally cautious proposal, not to a vetoed strong action that has just been downgraded. Preserving that distinction prevents the conceptual explanation from silently changing the reported executor into a different iterative decision procedure. The conservative Layer~2 policy text is reproduced in Appendix~\ref{app:prompts:rules}; it complements, but does not replace, the Layer~3 checks.

\subsection{A Worked Decision Walkthrough}
\label{sec:walkthrough}

Consider the two illustrative statements from the Introduction. ``I am not sure the second read is reliable'' expresses a data-quality judgment. Its tentative wording should remain visible in the source-linked representation, and it does not by itself authorize exclusion, truncation, or a threshold change. ``This part is invalid, truncate it'' expresses a firm directive, but executing a parameterized truncation still requires the relevant file scope, an existing boundary, and the required signal support. Firm language does not supply a timestamp that the record lacks.

\begin{table}[htbp]
\centering\small
\begin{tabularx}{\linewidth}{@{}>{\raggedright\arraybackslash}p{0.19\linewidth}YY@{}}
\toprule
\textbf{Decision step} & \textbf{Tentative quality concern} & \textbf{Firm truncation directive}\\
\midrule
Preserve the source & Keep the hedge and \judgment{} label. & Keep the explicit directive and \fact{} label.\\
Retain available structure & Preserve recorded scope and any already supplied candidate. & Preserve recorded scope, boundary, and authorization support.\\
Check the current file & Uncertainty alone is insufficient for strong intervention. & Check the action, existing parameter, and required signal evidence.\\
Handle insufficient support & Default to review-preserving handling. & Apply the appropriate fail-closed transition.\\
Corroborated candidate & Only the narrow, unique-candidate elevation rule may apply. & A cautious proposal may similarly be replaced only when that rule is satisfied.\\
\bottomrule
\end{tabularx}
\caption{Illustrative walkthrough, not additional experimental cases or literal capsule dumps. The example separates preserving a source directive from authorizing a parameterized action on a file.}
\label{tab:walkthrough}
\end{table}

To make the candidate case explicit, suppose a source-linked record already contains a truncation candidate with timestamp $t^\star$. If an originally cautious proposal is accompanied by exactly one eligible candidate whose stored timestamp matches the calibrated step drop, \textsc{Substitute} may output \field{TRUNCATE\_AT} with that same $t^\star$. Without the stored parameter, the required evidence match, or uniqueness, this substitution is unavailable. The symbol denotes an existing parameter; it is not a new measured value. The original source certainty remains unchanged in either outcome.

\section{Datasets and Evaluation Protocol}
\label{sec:datasets}\label{sec:setup}

The study contains two predictive evaluations and one deterministic compilation audit. Exp~1 evaluates Stage~1 directive extraction across three corpora. Exp~2 checks whether Stage~2 preserves fixed EDE records. Exp~3 evaluates downstream skill loading on three aligned FreeNotes sessions, with additional analyses of a certainty-only policy and model-predicted EDE inputs. The technical report reorganizes and explains these existing evaluations; it does not add new model runs or experimental sessions.

\subsection{Corpora, Alignment, and Units of Evaluation}
\label{sec:corpora}

\textbf{FreeNotes} contains 201 annotated segments from experimental notebooks contributed by two senior researchers at two institutions. The notebooks predate \systemname{}; each session is single-authored, written close to experimentation, and uses Chinese--English code-switched text. \textbf{ONS} contains 155 segments from nine entries on \field{openlabnotebooks.org}, under a CC BY license, and serves as a semi-formal boundary case. \textbf{WLP} contributes 105 prescriptive segments from the Wet Lab Protocols corpus~\citep{wlp_2018}, providing a high-formality protocol control.

\begin{table}[htbp]
\centering
\small
\setlength{\tabcolsep}{5pt}
\renewcommand{\arraystretch}{1.12}
\begin{tabular}{@{}p{0.23\columnwidth}r p{0.25\columnwidth} p{0.30\columnwidth}@{}}
\toprule
\textbf{Corpus} & \textbf{$n$} & \textbf{Formality} & \textbf{Role} \\
\midrule
FreeNotes & 201 & Low, bilingual
& In-regime; Exp~1--3 \\
ONS & 155 & Intermediate
& Boundary case; Exp~1--2 \\
WLP & 105 & High, prescriptive
& Protocol control; Exp~1--2 \\
\bottomrule
\end{tabular}
\caption{The three corpora used in this study. $n$ is the number of annotated segments. FreeNotes is the only corpus used for downstream evaluation because it is the only one with aligned raw instrument data and expert-adjudicated file labels.}
\label{tab:corpora}
\end{table}

Only FreeNotes supports the downstream evaluation, because its selected sessions align notebook directives, raw instrument records, and expert-adjudicated file-level decisions. There are 48 downstream files: 17 in uncertainty-heavy \emph{Saturation-A}, 22 in terminal \emph{Saturation-B}, and nine in \fact{}-dominated \emph{Step-drop}. ONS and WLP broaden the extraction and compilation settings, but they do not provide additional downstream instrument-data sessions.

\begin{figure}[tbp]
\centering
\includegraphics[width=0.68\linewidth]{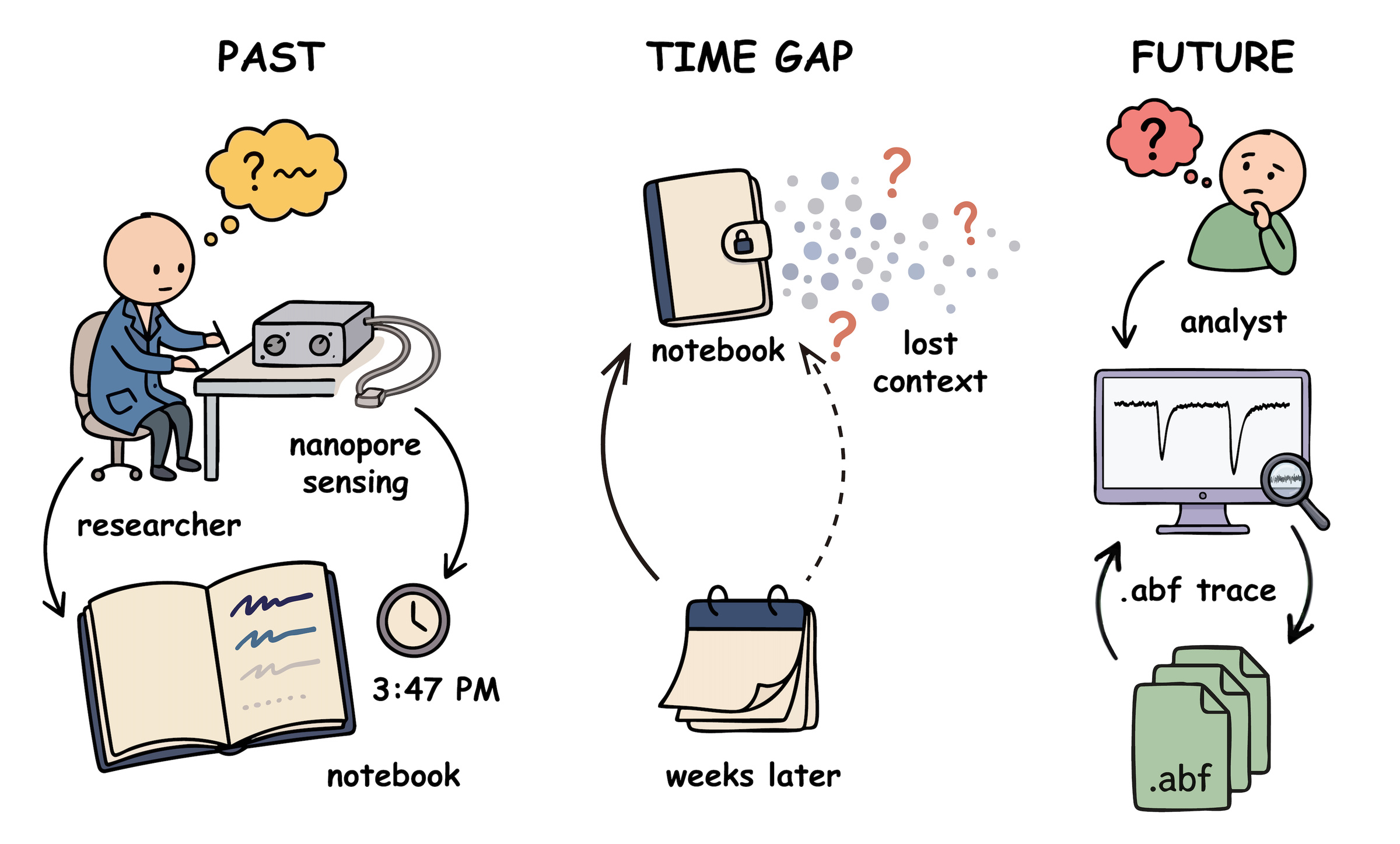}
\caption{Downstream validation setting. Notebook context links an experiment to later file-level data-handling decisions. The illustration motivates source retention across the time gap; the evaluated outputs are processing decisions over the aligned recordings.}
\label{fig:closed-loop-gap}
\end{figure}

The reported totals refer to different units. The 461 items are annotated text segments, the 149 items are fixed directive records in the compilation audit, and the 48 items are instrument files in the downstream evaluation. The predicted-EDE stress test extracts directives from 77 segments associated with the downstream sessions. Neither an audited capsule nor a repeated API call constitutes another independent wet-lab experiment. Keeping these units separate is necessary for interpreting the scope of the results.

\subsection{Annotation and Gold Construction}
\label{app:guideline}\label{app:formality}

All corpora use the shared guideline for directive presence, five directive types, and three epistemic statuses, with corpus-specific boundary examples. Two annotators independently labeled stratified subsamples: 60 FreeNotes segments, 60 ONS segments, and 90 WLP segments. The second annotator received the raw text, guideline, and a fixed exemplar set, without primary labels or LLM assistance. The minimum agreement across corpora is $\kappa=0.709$ for directive detection and QWK$=0.732$ for ordinal certainty.

\begin{table}[htbp]
\centering
\small
\setlength{\tabcolsep}{5pt}
\renewcommand{\arraystretch}{1.12}
\begin{tabular*}{\columnwidth}{@{\extracolsep{\fill}}lrr@{}}
\toprule
\textbf{Corpus} & \textbf{Annotated} & \textbf{Double-annotated} \\
\midrule
FreeNotes & 201 & 60 \\
ONS       & 155 & 60 \\
WLP       & 105 & 90 \\
\bottomrule
\end{tabular*}
\caption{Stage~1 annotation coverage. The double-annotated subsets were labeled independently by two annotators.}
\label{tab:annotation_coverage}
\end{table}

Corpus formality is characterized with three descriptive proxies: surface regularity, code-switching, and author--audience distance. FreeNotes is informal, self-directed, and frequently bilingual; ONS is predominantly English and reflective; WLP is predominantly English with regular imperative or template-like protocol language. The reported ordering WLP $>$ ONS $>$ FreeNotes holds on at least two of the three proxies. These descriptors explain the corpus selection; they are not an additional model input or a single calibrated formality score.

For Exp~3, gold actions were constructed from notebook directives, Layer~1 signal findings, and session metadata. Two annotators independently cross-checked the five-action labels, with disagreements adjudicated before agent outputs were evaluated. Gold actions and adjudication rationales do not enter prompts, capsules, or executor inputs. These are expert-adjudicated operational decisions grounded jointly in the available sources, not labels derived from the agent's proposed actions.

\subsection{Stage~1 Models, Splits, and Metrics}
\label{sec:setup:stage1}

Stage~1 evaluates GPT-4o, Claude Sonnet~4.5, and Qwen-Max under zero-shot and few-shot prompting at temperature~0. The reported setup uses document-stratified 80/20 splits and a shared test set of 87 segments: 31 FreeNotes, 21 WLP, and 35 ONS. One segment is supplied per call. Five of 522 calls remained empty after retry and are treated as parse failures; primary scores are computed over valid responses. Appendix~\ref{app:exp1_full} reports the coverage and per-corpus results.

Few-shot exemplars are sampled once from the document-stratified training fold with seed~42, using one example per observed directive type and one negative example. \field{CONDITION\_CHANGE} appears only in the FreeNotes training fold, yielding $k=6$ for FreeNotes and $k=5$ for WLP and ONS. The same exemplar set is reused for all test segments within a corpus. Complete documented prompt templates, including their placeholders, appear in Appendix~\ref{app:prompts}.

Binary $F_1^{\mathrm{hd}}$ measures directive detection. $F_1^{\mathrm{dt}*}$ and $F_1^{\mathrm{ep}*}$ are observed-label macro-$F_1$ scores on the \emph{both-positive subset}: segments for which both gold and prediction contain a directive. They assess type and epistemic status conditional on successful positive detection; they are not full-pipeline label scores over all segments. \emph{Joint} is the valid-response fraction for which all three labels match exactly.

Because the three statuses are ordered by downstream commitment in this framework, the evaluation also reports quadratic-weighted kappa~\citep{cohen_1968}:
\begin{equation}
 \kappa_Q=1-\frac{\sum_{i,j}w_{ij}O_{ij}}{\sum_{i,j}w_{ij}E_{ij}},
 \qquad w_{ij}=\frac{(i-j)^2}{(C-1)^2},\quad C=3.
 \label{eq:qwk}
\end{equation}
Here $O$ is the observed confusion matrix and $E$ the expected matrix under independence. The weights penalize \fact{}--\suggestion{} confusions more than adjacent-status confusions. This statistical independence is the chance-agreement construction for the metric, not an assumption that notebook and signal evidence are independent.

\subsection{Stage~2 Preservation Audit}
\label{sec:setup:stage2}

Exp~2 uses fixed EDE records and checks directive identity, unchanged certainty, source fidelity, and schema validity as described in Section~\ref{sec:preservation_contract}. FreeNotes is compiled per experimental session; WLP and ONS are compiled at corpus level for this audit. FreeNotes additionally uses the action-policy layer. Because this stage is deterministic, the audit evaluates preservation rather than predictive model quality.

\subsection{Downstream Conditions and Controlled Differences}
\label{sec:setup:conditions}

The seven original downstream configurations use the same model, five-action vocabulary, and per-file \field{SignalFindings}. Table~\ref{tab:conditions} distinguishes representation and policy-stack changes. Each condition--file pair is sampled through five separate API calls.

\begin{table}[htbp]
\centering\small
\renewcommand{\arraystretch}{1.12}
\begin{tabularx}{\linewidth}{@{}p{0.31\linewidth}Yp{0.14\linewidth}p{0.13\linewidth}@{}}
\toprule
\textbf{Condition} & \textbf{Representation} & \textbf{L2 policy block} & \textbf{L3 checking}\\
\midrule
External LLM (no skill) & Raw notebook and findings; standalone prompt & No & Off\\
Action-only skill & Flat actions and certainty metadata & No & Off\\
\quad + executor & Flat actions and certainty metadata & Yes & On\\
Raw notes & Source excerpts & No & Off\\
\systemname{} skill & Full \metaskill{} & No & Off\\
\quad + verify & Full \metaskill{} & Yes & A+V\\
\textbf{\quad + verify+elevate} & \textbf{Full \metaskill{}} & \textbf{Yes} & \textbf{A+V+S}\\
\bottomrule
\end{tabularx}
\caption{Seven original downstream configurations. A, V, and S denote Authorize, Veto, and Substitute. The proposed configuration is bold. The conservative Layer~2 block is added in every executor-enabled condition; the six non-external conditions share a common base prompt. The post-hoc Label-policy Gate is analyzed separately in Section~\ref{sec:label_policy}.}
\label{tab:conditions}
\end{table}

Across the six non-external conditions, a common base decision prompt and per-file evidence summary are held fixed, while suffixes and loaded representations differ. Executor-enabled conditions additionally receive the same conservative Layer~2 execution-policy block and deterministic Layer~3 checking. On/off comparisons therefore measure a combined policy-stack intervention, not an executor-only causal effect. The external baseline uses a standalone raw-notebook prompt and is likewise not a one-factor ablation of compilation.

The Action-only representation is not certainty-free. Its prompt exposes \field{epistemic\_status}, \field{confidence\_weight}, verbatim uncertainty markers, the raw excerpt, and a source label. It lacks the explicit authorization, default-action tiers, and candidate-action structures consumed by the full executor. Its suffix also resolves multiple applicable directives by choosing the most severe action. These representation and prompting differences must be considered when attributing performance changes; the comparison is not simply ``labels absent'' versus ``labels present''.

The post-hoc \emph{Label-policy Gate} transforms fixed Action-only proposals using the same epistemic labels and findings, but without full \metaskill{} authorization or candidate structures, executor decisions, or gold labels. Applicable judgments are routed to review, suggestions remain advisory, and strong actions require a matching fact and separately supported parameters. It makes no new Layer~2 calls. The predicted-EDE stress test instead replaces adjudicated extraction inputs in the proposed configuration with Claude Sonnet~4.5 few-shot predictions. These two analyses answer different questions and are reported separately.

\subsection{Downstream Metrics and Statistical Scope}
\label{sec:downstream_metrics}

For each condition--file pair, evaluation aggregates the five returned actions by the reported majority-vote procedure. Accuracy (Acc), balanced accuracy (bAcc), unweighted Cohen's $\kappa$, and \field{FLAG\_FOR\_REVIEW} recall (FR) summarize file-level agreement. Gold-support macro-$F_1$ (g$F_1$) averages class $F_1$ only over action labels present in that session's gold set; it is not a fixed five-label macro-average. Full per-session values are retained in Appendix~\ref{app:additional_downstream}.

The primary endpoint is agreement on discrete actions, so unweighted $\kappa$ is primary despite the severity ordering used in the prompt. Quadratic-weighted $\kappa$ is a sensitivity check. Action unanimity measures the fraction of files for which all five calls return the same action. The confidence intervals are computed over file-majority accuracy, not by treating repeated API calls as additional files. With only three sessions, these within-session estimates do not establish population-level robustness across laboratories or experimental domains.

\section{Results and Failure Analysis}
\label{sec:results}

The results distinguish three questions: whether models extract the intended directives, whether compilation preserves fixed input records, and whether the resulting skill and policy stack support appropriate downstream decisions. These are complementary forms of evidence rather than successive measurements of the same accuracy. In particular, perfect preservation of fixed EDE records should not be read as perfect end-to-end extraction.

\subsection{Exp~1: Extracting Directives and Author Certainty}
\label{sec:res:exp1}

Table~\ref{tab:exp1_main} reports pooled valid-response metrics on the shared 87-segment test set. Few-shot prompting increases directive-type macro-$F_1$ for all three backbones: Claude improves from 0.320 to 0.500, GPT-4o from 0.272 to 0.392, and Qwen-Max from 0.376 to 0.500. Claude few-shot attains the highest directive-detection $F_1$ (0.737) and joint accuracy (0.523), while GPT-4o zero-shot attains the highest certainty QWK (0.946). Better type detection and better certainty agreement do not necessarily occur under the same configuration.

\begin{table}[t]
\centering
\small
\renewcommand{\arraystretch}{1.15}
\setlength{\tabcolsep}{4pt}

\begin{tabular*}{\textwidth}{
  @{\extracolsep{\fill}}
  l
  l
  c
  c
  c
  c
  c
  @{}
}
\toprule
\textbf{Model}
& \textbf{Condition}
& $F_1^{\mathrm{hd}}$
& $F_1^{\mathrm{dt}*}$
& $F_1^{\mathrm{ep}*}$
& $\mathrm{QWK}^{\mathrm{ep}}$
& \textbf{Joint} \\
\midrule

Claude Sonnet~4.5
& zero-shot
& 0.682
& 0.320
& 0.838
& 0.781
& 0.470 \\

Claude Sonnet~4.5
& few-shot
& \textbf{0.737}
& \textbf{0.500}
& 0.830
& 0.761
& \textbf{0.523} \\

\addlinespace[2pt]

GPT-4o
& zero-shot
& 0.632
& 0.272
& \textbf{0.891}
& \textbf{0.946}
& 0.471 \\

GPT-4o
& few-shot
& 0.706
& 0.392
& 0.791
& 0.720
& 0.494 \\

\addlinespace[2pt]

Qwen-Max
& zero-shot
& 0.597
& 0.376
& 0.764
& 0.755
& 0.471 \\

Qwen-Max
& few-shot
& 0.711
& \textbf{0.500}
& 0.813
& 0.834
& 0.494 \\

\bottomrule
\end{tabular*}

\caption{
Exp.~1 pooled valid-response metrics on the shared 87-segment test set.
$F_1^{\mathrm{hd}}$ is binary directive-detection $F_1$;
$F_1^{\mathrm{dt}*}$ and $F_1^{\mathrm{ep}*}$ are observed-label
macro-$F_1$ scores on the both-positive subset;
\emph{Joint} is the three-tuple exact-match rate.
Bold indicates the best value in each metric column.
Parsing coverage ranges from 83/87 to 87/87;
full per-corpus results and parse statistics are reported in Appendix~A.
}
\label{tab:exp1_main}
\end{table}

The per-corpus results in Appendix~\ref{app:exp1_full} localize important variation. WLP is the most stable corpus, while FreeNotes shows substantial few-shot gains in type classification. On ONS, Qwen-Max's directive-detection $F_1$ rises from 0.154 to 0.727 with exemplars. This supports the value of exemplar grounding in applying the task definition, but does not establish that the remaining extraction problem is solved. Some class-level scores also rest on very small both-positive subsets: three examples for GPT-4o zero-shot on ONS and one for Qwen-Max zero-shot.

Certainty errors concentrate at the \fact{}--\judgment{} boundary, with eight reported errors in one direction and five in the reverse. This is a consequential distinction for downstream commitment, even when an action category is otherwise plausible. More generally, both-positive label metrics omit false-positive and false-negative directive detections. Joint accuracy and the predicted-input experiment are therefore needed to understand the limitations of the full pipeline.

\subsection{Exp~2: Preserving Fixed Directives during Compilation}
\label{sec:res:exp2}\label{app:stage2_audit}

The deterministic compiler emits all 149 fixed input directives: 48 from FreeNotes, 70 from WLP, and 31 from ONS. Every capsule passes the four universal checks, and all 48 FreeNotes capsules pass the additional policy check (Table~\ref{tab:exp2}). The same paper-level structure can therefore be inspected across bilingual notebooks, semi-formal entries, and prescriptive protocols.

\begin{table}[htbp]
\centering
\small
\setlength{\tabcolsep}{5pt}
\renewcommand{\arraystretch}{1.12}
\begin{tabular*}{\columnwidth}{@{\extracolsep{\fill}}lrrrr@{}}
\toprule
\textbf{Corpus} & \textbf{EDE} & \textbf{Caps.} & \textbf{All checks} & \textbf{Policy} \\
\midrule
FreeNotes & 48 & 48 & 48/48 & 48/48 \\
WLP       & 70 & 70 & 70/70 & -- \\
ONS       & 31 & 31 & 31/31 & -- \\
\bottomrule
\end{tabular*}
\caption{Cross-corpus Stage~2 preservation audit. ``All checks'' aggregates the four universal checks for fixed input directives. A dash denotes that the downstream action-policy layer is not used for that corpus.}
\label{tab:exp2}
\end{table}

This finding establishes fidelity to the fixed compilation records, not an extraction recall of 100\% over all notebook text. It also does not demonstrate downstream execution in ONS or WLP. The audit's value is a checkable separation: errors introduced by a predictive extractor should not be confused with directives lost or altered during deterministic compilation.

\subsection{Exp~3: Downstream Decisions across Epistemic Regimes}
\label{sec:res:exp3}

Table~\ref{tab:downstream} summarizes the original seven-condition comparison on the 48 nanopore files, using the fixed/adjudicated EDE inputs where compiled directives are required. The two saturation sessions expose failures that are not apparent on the \fact{}-dominated Step-drop session. The external and Action-only baselines have review recall of 0--21.4\% and near-chance $\kappa$ on the saturation sessions, but both reach $\kappa=0.80$ on Step-drop.

\begin{table}[htbp]
\centering
\small
\setlength{\tabcolsep}{5pt}
\renewcommand{\arraystretch}{1.12}
\begin{tabular*}{\columnwidth}{@{\extracolsep{\fill}}l rr rr rr@{}}
\toprule
& \multicolumn{2}{c}{\textit{Sat-A}}
& \multicolumn{2}{c}{\textit{Sat-B}}
& \multicolumn{2}{c}{\textit{Step}} \\
\cmidrule(lr){2-3}\cmidrule(lr){4-5}\cmidrule(l){6-7}
\textbf{Condition}
& $\kappa$ & FR & $\kappa$ & FR & $\kappa$ & FR \\
\midrule
External LLM
& $-0.05$ & 0.0 & 0.09 & 0.0 & 0.80 & 100.0 \\
Action-only
& 0.14 & 21.4 & 0.09 & 0.0 & 0.80 & 100.0 \\
\quad + executor
& 0.47 & 100.0 & 0.00 & 100.0 & 0.00 & 100.0 \\
Raw notes
& 0.17 & 50.0 & 0.09 & 0.0 & 0.63 & 75.0 \\
\systemname{} skill
& 0.13 & 0.0 & 0.14 & 30.0 & 0.15 & 100.0 \\
\quad + verify
& 0.71 & 85.7 & 1.00 & 100.0 & 0.15 & 100.0 \\
\textbf{\quad + verify+elevate}
& \textbf{0.71} & \textbf{85.7}
& \textbf{1.00} & \textbf{100.0}
& \textbf{0.63} & \textbf{100.0} \\
\bottomrule
\end{tabular*}
\caption{Main downstream ablation. $\kappa$ is unweighted and FR is \texttt{FLAG\_FOR\_REVIEW} recall (\%). The main comparison uses fixed/adjudicated EDE inputs for compiled skills. Full per-session metrics and additional checks are in Appendix~\ref{app:additional_downstream}; Label-policy and predicted-EDE results are in Sections~\ref{sec:label_policy} and~\ref{sec:predicted_ede}. Bold identifies the proposed configuration, not a best-in-column claim.}
\label{tab:downstream}
\end{table}

With a full \metaskill{}, the verify policy stack raises review recall to 85.7\% on Saturation-A and 100\% on Saturation-B, while balanced accuracy reaches 96.4\% and 100\%, respectively. The corresponding Action-only+executor condition instead collapses to always-review on Saturation-B and Step-drop. This contrast shows why a gate needs the authorization and candidate structure it is designed to inspect; attaching it to a representation lacking those fields is not equivalent to supplying the full contract.

Candidate elevation affects only Step-drop in the reported comparison. Verify and verify+elevate are identical on the saturation sessions, with no elevation events. On Step-drop, Substitute is active in 14 of 45 calls. The reported file-majority accuracy increases from 44.4\% to 77.8\%, and $\kappa$ from 0.15 to 0.63. The proposed configuration does not outperform every baseline in every session: Action-only remains better on Step-drop. Its contribution is avoiding both observed regime-level failures within this evaluation, rather than universal dominance.

\subsection{Structured Authorization versus Certainty Labels Alone}
\label{sec:label_policy}

The Label-policy Gate provides a more informative comparison than treating Action-only as a baseline with no uncertainty information. It reuses the fixed Action-only proposals and applies a deterministic certainty-aware policy, without the \metaskill{} authorization and candidate structures. Table~\ref{tab:label_policy} retains the full reported results and distinguishes this post-hoc intervention from the original configurations.

\begin{table}[htbp]
\centering
\small
\setlength{\tabcolsep}{5pt}
\renewcommand{\arraystretch}{1.12}
\begin{tabular*}{\columnwidth}{@{\extracolsep{\fill}}llrrrrr@{}}
\toprule
\textbf{Session} & \textbf{Method} & Acc & bAcc & g$F_1$ & $\kappa$ & FR \\
\midrule
\multirow{3}{*}{Sat-A} & Action-only & 29.4 & 55.4 & 39.4 & .139 & 21.4 \\
 & Label-policy & 82.4 & 48.2 & 47.4 & .338 & 92.9 \\
 & Full & 88.2 & 96.4 & 85.6 & .707 & 85.7 \\
\midrule
\multirow{3}{*}{Sat-B} & Action-only & 9.1 & 66.7 & 66.7 & .087 & 0.0 \\
 & Label-policy & 54.5 & 51.7 & 56.2 & .095 & 55.0 \\
 & Full & 100.0 & 100.0 & 100.0 & 1.000 & 100.0 \\
\midrule
\multirow{3}{*}{Step} & Action-only & 88.9 & 66.7 & 63.0 & .800 & 100.0 \\
 & Label-policy & 88.9 & 66.7 & 63.0 & .800 & 100.0 \\
 & Full & 77.8 & 58.3 & 58.2 & .633 & 100.0 \\
\bottomrule
\end{tabular*}
\caption{Label-policy Gate applied to fixed Action-only proposals. ``Full'' denotes \systemname{} with verify+elevate.}
\label{tab:label_policy}
\end{table}

On Saturation-A, Label-policy achieves higher review recall than the full configuration (92.9\% versus 85.7\%), but markedly lower balanced accuracy (48.2\% versus 96.4\%). On Saturation-B, it improves review recall over Action-only but remains well below the full configuration in both balanced accuracy and agreement. On Step-drop, Label-policy leaves the stronger Action-only result unchanged. Thus, labels support a useful conservative policy, but review recall alone does not establish balanced recovery of the session's actions.

This analysis supports the practical role of structured authorization in the tested stack, not a claim that no other certainty-aware gate could work. The full and Action-only pipelines differ in representation and prompting, while Label-policy is applied post hoc to fixed proposals. A matched-policy experiment would be required to isolate every contribution of the representation, prompt, and executor.

\subsection{Error Propagation from Model-Predicted EDE}
\label{sec:predicted_ede}

The primary downstream comparison conditions on fixed/adjudicated EDE inputs. Replacing them with Claude Sonnet~4.5 few-shot predictions asks a separate end-to-end question. Across Saturation-A, Saturation-B, and Step-drop, directive recall is 87.5\%, 100\%, and 92.3\%, but precision is 87.5\%, 34.8\%, and 47.9\%, respectively. High recall does not protect the downstream system from over-detected directives.

\begin{table}[htbp]
\centering\small
\renewcommand{\arraystretch}{1.12}
\begin{tabularx}{\linewidth}{@{}p{0.21\linewidth}Yrrrr@{}}
\toprule
\textbf{Session} & \textbf{EDE input} & \textbf{Acc} & \textbf{bAcc} & \textbf{g$F_1$} & $\boldsymbol\kappa$\\
\midrule
Saturation-A & Adjudicated & 88.2 & 96.4 & 85.6 & 0.71\\
 & Predicted & 82.4 & 71.4 & 59.7 & 0.51\\
\addlinespace
Saturation-B & Adjudicated & 100.0 & 100.0 & 100.0 & 1.00\\
 & Predicted & 68.2 & 25.0 & 27.0 & $-0.07$\\
\addlinespace
Step-drop & Adjudicated & 77.8 & 58.3 & 58.2 & 0.63\\
 & Predicted & 22.2 & 16.7 & 14.8 & $-0.03$\\
\bottomrule
\end{tabularx}
\caption{Adjudicated versus model-predicted EDE in the full verify+elevate configuration. Both rows use results already reported in the camera-ready manuscript. Predicted EDE is obtained from 77 downstream-session segments. Acc, bAcc, and g$F_1$ are percentages; $\kappa$ is unweighted.}
\label{tab:predicted_ede}
\end{table}

Table~\ref{tab:predicted_ede} makes the gap explicit. Agreement degrades modestly on Saturation-A but falls below chance on Saturation-B and Step-drop. In this stress test, Stage~1 precision is the main observed bottleneck. Faithful compilation can preserve an incorrectly extracted directive, and evidence checks cannot be assumed to recover the scientist's intended guidance when the supplied record is wrong. The finding limits claims of autonomous end-to-end reliability despite the strong fixed-input results.

The table measures action agreement, not a separate rate of unsupported strong actions. A drop in $\kappa$ cannot by itself show either that safety constraints were violated or that they remained perfectly satisfied. Such a claim would require the relevant action-level authorization and evidence audit. This distinction keeps extraction quality, preservation fidelity, and execution behavior analytically separate.

\subsection{Small Samples, Repeated Calls, and Baseline Collapse}
\label{sec:result_robustness}

The Wilson 95\% intervals for full-configuration accuracy are [65.7, 96.7] on Saturation-A (15/17), [85.1, 100.0] on Saturation-B (22/22), and [45.3, 93.7] on Step-drop (7/9). They reflect the small number of files, not the five repeated API calls per file. In particular, perfect observed accuracy in one 22-file session is not a guarantee of performance on a new session.

Repeated-call consistency is also different from correctness. On Saturation-B, the external baseline, Action-only+executor, and full configuration all have action unanimity 1.00, but their $\kappa$ values are 0.09, 0.00, and 1.00. Stable output can therefore represent stable error. Similarly, the always-review baseline has perfect review recall by construction while missing the other required actions. Appendix~\ref{app:additional_downstream} retains these checks and the quadratic-weighting sensitivity analysis. The results support a regime-specific mechanism interpretation, not a large-sample claim of general scientific-agent reliability.

\section{Related Work and Positioning}
\label{sec:related}

\systemname{} connects scientific procedural-text extraction, source uncertainty, and agent skill construction. Its distinctive setting is a single-author notebook whose qualifications must survive the transition from text to action. The following connections situate that problem without treating every uncertainty-bearing document or every reusable agent artifact as an instance of the same task.

\subsection{Scientific Procedure Extraction and Protocol Translation}

Procedural-text extraction has developed representations of actions and their relations across several document genres. Recipe flow graphs~\citep{mori_recipe_2014}, the Wet Lab Protocols corpus~\citep{wlp_2018}, and X-WLP's process-level representation~\citep{xwlp_2021} provide important foundations. BioPlanner evaluates LLM-based protocol planning in biology~\citep{bioplanner_2023}, while NERRE addresses flexible-schema entity and relation extraction from scientific text~\citep{nerre_2024}. Protocol-to-DSL and protocol-translation systems further connect published procedures to machine-readable or executable representations~\citep{shi_protocol_2024,protocode_2024,cronin_xdl_2020}.

Our distinction is not that protocols never contain qualifiers: WLP is included precisely because the same label schema can be applied across a formality spectrum. Rather, experimental notebooks place observations, uncertain interpretations, and prospective actions in the same evolving account. A record may be useful for later analysis without being prescriptive. The notebook-to-skill task must preserve whether a source supports an action, not only identify the action or its arguments.

\subsection{Agent Skills, Workflows, and Memory}

SYNTACT resolves ambiguity in enterprise standard operating procedures through a Clarifier--Planner--Implementor dialogue~\citep{syntact_2025}. Flow-of-Action uses retrieved or generated SOPs to constrain multi-agent root-cause-analysis trajectories~\citep{flow_of_action_2025}. The Agent Skills format provides a practical way to package reusable instructions and associated resources~\citep{anthropic_skills_2025}. These are relevant precedents for turning procedural knowledge into artifacts an agent can load.

Agents also acquire reusable capabilities from interaction and experience, including embodied skill acquisition, continual task adaptation, and workflow memory~\citep{voyager_2023,clin_2023,awm_2025}. In \systemname{}, the primary knowledge source is instead the scientist's record. The representation keeps that source linked to each directive and makes its author certainty operational. Apparent ambiguity may express unresolved scientific judgment, in which case removing it is not an improvement in the skill.

This source choice also distinguishes the present work from an evaluation of continual agent learning. The reported experiments do not optimize a skill library through successive environmental rewards, update memory after each trial, or measure long-term skill evolution. They evaluate extraction, deterministic compilation, and use of fixed notebook-derived skills. Such skills could be components of a larger memory or workflow system, but that integration is not an experimental result here.

\subsection{Tacit Knowledge and Records of Scientific Practice}

Polanyi and Collins emphasize aspects of knowledge and scientific practice that are not exhausted by formal descriptions~\citep{polanyi1966tacit,collins2001tacit}. Notebook research examines how written and multimodal records participate in knowledge-making activity~\citep{doody2022labbook}. These perspectives motivate attention to the experimental process rather than only polished publications.

Our operational target is narrower than complete tacit-knowledge extraction. It is the portion expressed in notebook text: a stated concern, a qualification, a practical judgment, or a next-step proposal that can affect subsequent processing. Source certainty is part of what must be retained from that record. Section~\ref{sec:tacit} makes this boundary explicit so that success on directive extraction is not mistaken for recovering unrecorded expertise.

\subsection{Uncertainty, Factuality, and Action Authorization}

Author uncertainty and factuality have a long history in NLP. BioScope and the CoNLL-2010 shared task annotate uncertainty, hedges, and their scope~\citep{bioscope_2008,conll_2010}. Work on modality, FactBank, and ORCA models further dimensions of factuality, certainty, and attribution~\citep{thompson_modality_2008,factbank_2009,orca_2012}. Clinical assertion classification and the CMED schema associate events with uncertainty and other contextual attributes~\citep{i2b2_2010,n2c2_cmed_2022}.

EDE builds on the need to represent what a source commits to, but uses that information for a downstream decision interface. Its epistemic label is not merely descriptive metadata: the label, source scope, authorization fields, and signal evidence jointly constrain action. The three-status schema is a task-specific operational choice, not a replacement for richer factuality or modality annotation. The predicted-EDE results show why the quality of this upstream linguistic analysis remains consequential even with deterministic compilation.

\subsection{Scientific Agents and Autonomous Laboratories}

Scientific-agent systems have demonstrated tool-assisted chemistry, autonomous chemical research, and integrated laboratory workflows~\citep{coscientist_2023,chemcrow_2024,alab_2023}. Co-Scientist supports scientific hypothesis generation and research planning~\citep{gottweis_coscientist_2026}. These systems motivate the broader question of what scientific knowledge an agent can use and how that knowledge is represented.

\systemname{} addresses one component of that broader agenda: preserving the scientist's stated certainty when experimental notes become reusable guidance. Its nanopore evaluation concerns later file-handling decisions, not replacement of the complete scientific workflow. The results therefore support an evidence-aware notebook-to-decision interface rather than a claim of general autonomous scientific discovery.

\section{Portability and Reproducibility}
\label{sec:reproducibility}

The reusable part of \systemname{} is an epistemic contract, not a universal processing policy. Directive labels, source links, and preservation checks can remain stable while the actions, parameters, evidence, and fallback rules are adapted to a domain. Portability of this interface must be distinguished from demonstrated transfer of downstream performance.

\subsection{What Must Be Specified for a New Domain}

Deployment requires the components of Equation~\ref{eq:adapter} and a mapping from source segments and their scope to the object being processed. Without these relationships, a certainty label cannot establish whether a directive applies to a particular file.

A new domain must specify its own evidence requirements and fallback actions rather than inherit the nanopore truncation rule. FreeNotes, ONS, and WLP support the use of a common extraction and compilation schema across writing regimes; they do not validate execution outside nanopore.

\subsection{Three Levels of Reproduction}
\label{sec:reproduction_levels}

\paragraph{Reported-table reproduction.}
The study's described research artifact includes derived \field{SignalFindings}, fixed five-call decision records, adjudicated file-level labels, evaluation scripts, and the Label-policy implementation. These records support reproduction of the reported downstream tables, post-hoc policy comparison, and repeated-call analyses without obtaining new model outputs. This reproduces the recorded evaluation, not a new independent experiment.

\paragraph{Extraction and agent reruns.}
The documented prompts, model choices, decoding settings, train-fold exemplar construction, and response-filtering policy specify the experimental setup. New API calls constitute new runs and should be distinguished from replaying the recorded decisions. The exact templates reproduced in Appendix~\ref{app:prompts} include runtime placeholders; they are not the complete instantiated messages for every file. The original UTF-8 bilingual strings and per-example inputs must be taken from the associated research materials where the templates refer to them.

\paragraph{Raw-signal reproduction.}
Recomputing Layer~1 from original ABF traces requires access to those traces. The raw recordings are not redistributed under the data-sharing agreement, whereas derived findings and the decision-level records are described as part of the released research artifact. Consequently, decision-level reproducibility and independent re-extraction of instrument evidence have different access requirements. Availability of the former does not imply public availability of the latter.

\subsection{Validation Boundaries and Further Evaluation}

Three extensions follow directly from the observed limitations. First, a matched-policy comparison should hold the agent's instructions constant when isolating the contribution of capsule structure and executor checking. The existing on/off comparison changes the policy stack as documented in Section~\ref{sec:setup:conditions}. Second, a new-session evaluation should distinguish development of the domain rules and calibration procedure from evaluation of those rules on independent recordings. Keeping gold labels out of runtime inputs is necessary, but is not by itself evidence of an independent rule-development split. Third, predicted-input evaluation should assess the richer source-linked record as well as the three EDE labels, since erroneous directives already impair the present end-to-end results.

These are further validation requirements, not additional experiments reported in this technical report. They preserve a clear boundary between what has been measured---cross-corpus extraction, fixed-record compilation, and three-session file-level decisions---and what remains to be established for broader scientific-agent use.

\section{Conclusion}
\label{sec:conclusion}

Laboratory notebooks expose a useful, partial record of scientific practice: observations, practical judgments, uncertainty, and possible next steps. \systemname{} turns this recorded guidance into agent-loadable skills while keeping each directive linked to its source and author certainty. EDE identifies the relevant statements; deterministic \metaskill{} compilation preserves fixed input records; a domain-specific policy stack checks processing actions against authorization and signal evidence.

The evaluation supports these components at different levels. All 149 fixed directives are preserved across three corpora. On 48 nanopore files, the full verify+elevate configuration avoids the two observed regime-level failures, while simpler alternatives can either over-commit on uncertain notes or miss firm directives. It is not the best configuration in every session, and model-predicted EDE causes substantial downstream degradation. Those limits are part of the result, not exceptions to it.

Preserving textual traces of tacit scientific knowledge requires a clear separation between source fidelity, scientific validity, and the evidence that authorizes an agent's action.

\section*{Limitations}
\phantomsection\addcontentsline{toc}{section}{Limitations}

This work studies lab notebooks with strong supporting context: the notes were written close to data collection, the raw records were preserved, and expert collaborators could judge how textual triage should affect later processing. This makes the evaluation more faithful than a standard crowd-sourced text task, but also harder to scale. Building FreeNotes required sustained collaboration with senior experimental researchers across two institutions, along with annotator training and repeated adjudication. Future work can broaden the scientific domains and operator backgrounds as more laboratories make notebooks, raw records, and expert judgments available under suitable data-sharing agreements.

The downstream evaluation is limited to three nanopore sessions and uses a single downstream model. The primary comparison conditions on fixed/adjudicated EDE inputs, and the predicted-input stress test shows that extraction errors remain a substantial limitation. The executor-enabled comparison also changes the Layer~2 policy block, so it cannot isolate an executor-only causal effect. Compilation fidelity does not independently validate scientific claims, infer entirely unrecorded expertise, or establish safe execution in untested domains.

\section*{Ethics Statement}
\phantomsection\addcontentsline{toc}{section}{Ethics Statement}

\systemname{} extracts structured directives from lab notebook text.
The FreeNotes corpus was contributed by two senior experimental researchers under a formal inter-institutional data-sharing agreement.
The notebooks were authored independently of the \systemname{} framework and used with the contributors' consent.
No third-party personal or sensitive data is present.
ONS entries are drawn from openlabnotebooks.org under their open-content license with author attribution preserved.
WLP segments are drawn from the Wet Lab Protocols corpus~\citep{wlp_2018} under its distribution terms.

The epistemic-grading mechanism preserves author uncertainty rather than replacing human judgment.
Downstream agent systems loading \metaskill{}s should treat epistemic labels and source provenance as inputs to review-preserving decision policies, not as authorizations to bypass human oversight.
Compiled skills are intended to support review-preserving scientific workflows, especially when downstream decisions affect experimental data inclusion or exclusion.

LLMs were used for Stage~1 extraction and downstream agent evaluation.
Stage~2 preservation artifacts are produced by deterministic compilers from fixed EDE inputs.
No LLM was used to create or adjudicate the reported gold labels; LLM outputs were used only in the evaluated extraction and agent conditions.
Where raw ABF traces cannot be redistributed under the data-sharing agreement, the released artifact includes anonymized FreeNotes segments, EDE labels, adjudicated file-level decisions, derived \texttt{SignalFindings}, fixed five-call decision records, the Label-policy implementation, compiler and executor code, evaluation scripts, tests, and reproduction instructions, together with the ONS/WLP portions under their original licenses. Appendix~\ref{app:prompts} summarizes prompt construction, and the exact templates used for the reported runs are reproduced in Appendix~\ref{app:prompts}. The released records reproduce the reported downstream tables and repeated-call analyses; the Layer~1 transformation from raw ABF traces to \texttt{SignalFindings} requires access to the original data (Section~\ref{sec:reproduction_levels}).

\section*{Acknowledgments}
\phantomsection\addcontentsline{toc}{section}{Acknowledgments}

This work was supported by the New Generation Artificial Intelligence--National Science and Technology Major Project (2026ZD0127802) and the Enterprise Ireland Commercialisation Fund (CF-2024-2410-F). We also thank our collaborators in Ireland for providing experimental data and supporting the data interpretation used in this study.

\clearpage

\bibliographystyle{plainnat}
\bibliography{references}

\clearpage
\appendix

\section{Full Stage~1 Results}
\label{app:exp1_full}

Tables~\ref{tab:exp1_freenotes}--\ref{tab:exp1_ons} report the per-corpus Stage~1 results corresponding to the pooled comparison in Table~\ref{tab:exp1_main}. WLP is the most stable corpus; FreeNotes shows the largest few-shot gains in directive-type classification; ONS is the most variable. The largest epistemic-status error flow is across the \texttt{FACT}--\texttt{JUDGMENT} boundary (8 \texttt{FACT}$\to$\texttt{JUDGMENT}; 5 in the reverse direction).

\begin{table}[htbp]
\centering
\small
\setlength{\tabcolsep}{5pt}
\renewcommand{\arraystretch}{1.12}
\begin{tabular*}{\columnwidth}{@{\extracolsep{\fill}}llccccc@{}}
\toprule
\textbf{Model} & \textbf{Cond.} & $F_1^{\rm hd}$ & $F_1^{\rm dt*}$ & $F_1^{\rm ep*}$ & QWK & Joint \\
\midrule
Claude   & zs & 0.611 & 0.242 & 0.861 & 0.788 & 0.267 \\
Claude   & fs & 0.714 & 0.500 & 0.930 & 0.953 & 0.419 \\
GPT-4o   & zs & 0.562 & 0.114 & 0.915 & 0.880 & 0.323 \\
GPT-4o   & fs & 0.667 & 0.345 & 0.709 & 0.719 & 0.355 \\
Qwen-Max & zs & 0.588 & 0.367 & 0.444 & 0.444 & 0.323 \\
Qwen-Max & fs & 0.667 & 0.583 & 0.852 & 0.914 & 0.452 \\
\bottomrule
\end{tabular*}
\caption{Stage~1 results on FreeNotes.}
\label{tab:exp1_freenotes}
\end{table}

\begin{table}[htbp]
\centering
\small
\setlength{\tabcolsep}{5pt}
\renewcommand{\arraystretch}{1.12}
\begin{tabular*}{\columnwidth}{@{\extracolsep{\fill}}llccccc@{}}
\toprule
\textbf{Model} & \textbf{Cond.} & $F_1^{\rm hd}$ & $F_1^{\rm dt*}$ & $F_1^{\rm ep*}$ & QWK & Joint \\
\midrule
Claude   & zs & 0.727 & 0.390 & 1.000$^\dagger$ & 1.000$^\dagger$ & 0.333 \\
Claude   & fs & 0.750 & 0.328 & 1.000$^\dagger$ & 1.000$^\dagger$ & 0.333 \\
GPT-4o   & zs & 0.800 & 0.320 & 1.000$^\dagger$ & 1.000$^\dagger$ & 0.333 \\
GPT-4o   & fs & 0.774 & 0.380 & 0.916 & 0.833 & 0.333 \\
Qwen-Max & zs & 0.800 & 0.413 & 0.916 & 0.833 & 0.381 \\
Qwen-Max & fs & 0.750 & 0.306 & 0.916 & 0.833 & 0.238 \\
\bottomrule
\end{tabular*}
\caption{Stage~1 results on WLP. $^\dagger$Epistemic-status metrics use the two labels present in the WLP test fold.}
\label{tab:exp1_wlp}
\end{table}

\begin{table}[htbp]
\centering
\small
\setlength{\tabcolsep}{5pt}
\renewcommand{\arraystretch}{1.12}
\begin{tabular*}{\columnwidth}{@{\extracolsep{\fill}}llccccc@{}}
\toprule
\textbf{Model} & \textbf{Cond.} & $F_1^{\rm hd}$ & $F_1^{\rm dt*}$ & $F_1^{\rm ep*}$ & QWK & Joint \\
\midrule
Claude   & zs & 0.737 & 0.339 & 0.650 & 0.364 & 0.750 \\
Claude   & fs & 0.762 & 0.563 & 0.333 & 0.071 & 0.735 \\
GPT-4o   & zs & 0.429 & 0.000$^\ddagger$ & 0.556$^\ddagger$ & 0.800$^\ddagger$ & 0.686 \\
GPT-4o   & fs & 0.667 & 0.190 & 0.667 & 0.400 & 0.714 \\
Qwen-Max & zs & 0.154 & 0.000$^\ddagger$ & 1.000$^\ddagger$ & 1.000$^\ddagger$ & 0.657 \\
Qwen-Max & fs & 0.727 & 0.409 & 0.556 & 0.739 & 0.686 \\
\bottomrule
\end{tabular*}
\caption{Stage~1 results on ONS. $^\ddagger$Both-positive $n$ is 3 for GPT-4o zero-shot and 1 for Qwen-Max zero-shot; the marked class-level values are therefore descriptive.}
\label{tab:exp1_ons}
\end{table}

Parse coverage is 30/31--31/31 on FreeNotes, 21/21 for all WLP cells, and 32/35--35/35 on ONS. Few-shot prompting shifts GPT-4o and Qwen-Max from more false negatives to more false positives in directive detection, while Claude remains false-positive dominant in both settings.

\subsection{Coverage and Interpretation}
Five of 522 Stage~1 calls remained empty after retry and are excluded from the primary valid-response scores. Each model--prompt configuration targets the same 87 segments, but its valid-response denominator can differ. A high conditional class score on a very small both-positive subset should not be interpreted as broad certainty-label reliability. The main evaluation protocol is given in Section~\ref{sec:setup:stage1}, and the documented prompts are in Appendix~\ref{app:prompts}.

\clearpage

\section{Full Downstream Results and Additional Checks}
\label{app:additional_downstream}

\subsection{Per-Session Metrics}
Table~\ref{tab:downstream_full} retains the complete per-session metrics for the original seven conditions. These compiled-skill comparisons use fixed/adjudicated EDE inputs. Acc, bAcc, g$F_1$, and FR are percentages; $\kappa$ is unweighted Cohen's $\kappa$. Bold identifies the proposed configuration rather than the best value in each column. Label-policy and predicted-EDE analyses appear in Sections~\ref{sec:label_policy} and~\ref{sec:predicted_ede}.

\begin{table}[htbp]
\centering\small
\setlength{\tabcolsep}{5pt}\renewcommand{\arraystretch}{1.07}
\begin{tabular*}{\linewidth}{@{\extracolsep{\fill}}lrrrrr@{}}
\toprule
\textbf{Condition} & \textbf{Acc} & \textbf{bAcc} & \textbf{g$F_1$} & $\boldsymbol\kappa$ & \textbf{FR}\\
\midrule
\multicolumn{6}{@{}l}{\textit{Saturation-A ($n=17$)}}\\[2pt]
External LLM & 0.0 & 0.0 & 0.0 & $-0.05$ & 0.0 \\
Action-only & 29.4 & 55.4 & 39.4 & 0.14 & 21.4 \\
\quad + executor & 88.2 & 50.0 & 48.3 & 0.47 & 100.0 \\
Raw notes & 47.1 & 37.5 & 33.3 & 0.17 & 50.0 \\
\systemname{} skill & 17.6 & 75.0 & 53.1 & 0.13 & 0.0 \\
\quad + verify & 88.2 & 96.4 & 85.6 & 0.71 & 85.7 \\
\textbf{\quad + verify+elevate} & \textbf{88.2} & \textbf{96.4} & \textbf{85.6} & \textbf{0.71} & \textbf{85.7} \\
\midrule
\multicolumn{6}{@{}l}{\textit{Saturation-B ($n=22$)}}\\[2pt]
External LLM & 9.1 & 66.7 & 66.7 & 0.09 & 0.0 \\
Action-only & 9.1 & 66.7 & 66.7 & 0.09 & 0.0 \\
\quad + executor & 90.9 & 33.3 & 31.7 & 0.00 & 100.0 \\
Raw notes & 9.1 & 66.7 & 66.7 & 0.09 & 0.0 \\
\systemname{} skill & 36.4 & 76.7 & 65.4 & 0.14 & 30.0 \\
\quad + verify & 100.0 & 100.0 & 100.0 & 1.00 & 100.0 \\
\textbf{\quad + verify+elevate} & \textbf{100.0} & \textbf{100.0} & \textbf{100.0} & \textbf{1.00} & \textbf{100.0} \\
\midrule
\multicolumn{6}{@{}l}{\textit{Step-drop ($n=9$)}}\\[2pt]
External LLM & 88.9 & 66.7 & 63.0 & 0.80 & 100.0 \\
Action-only & 88.9 & 66.7 & 63.0 & 0.80 & 100.0 \\
\quad + executor & 44.4 & 33.3 & 20.5 & 0.00 & 100.0 \\
Raw notes & 77.8 & 58.3 & 58.2 & 0.63 & 75.0 \\
\systemname{} skill & 44.4 & 33.3 & 24.2 & 0.15 & 100.0 \\
\quad + verify & 44.4 & 33.3 & 24.2 & 0.15 & 100.0 \\
\textbf{\quad + verify+elevate} & \textbf{77.8} & \textbf{58.3} & \textbf{58.2} & \textbf{0.63} & \textbf{100.0} \\
\bottomrule
\end{tabular*}
\caption{Full downstream results for all three sessions. Acc, bAcc, g$F_1$, and FR are percentages; $\kappa$ is unweighted. The full configuration is bold in each group.}
\label{tab:downstream_full}
\label{tab:downstream_satA}\label{tab:downstream_satB}\label{tab:downstream_step}
\end{table}

\subsection{Accuracy Uncertainty}
\label{app:downstream_ci}

Wilson 95\% confidence intervals for file-majority accuracy reflect the small number of evaluated files rather than the five repeated API calls. For full \systemname{}, the intervals are [65.7, 96.7] on Saturation-A (15/17), [85.1, 100.0] on Saturation-B (22/22), and [45.3, 93.7] on Step-drop (7/9). For Action-only, the corresponding intervals are [13.3, 53.1] (5/17), [2.5, 27.8] (2/22), and [56.5, 98.0] (8/9).

\subsection{Always-Review, Repeat Stability, and Weighting}
\label{app:always-flag}
\label{app:repeat_analysis}
\label{app:kappa_sensitivity}

A trivial always-review predictor has bAcc, g$F_1$, and $\kappa$ of (25.0, 22.6, 0.000) on Saturation-A, (33.3, 31.7, 0.000) on Saturation-B, and (33.3, 20.5, 0.000) on Step-drop. Repeated-call stability does not explain the main comparison: on Saturation-B, action unanimity is 1.00 for External LLM, Action-only+executor, and full \systemname{}, while their $\kappa$ values are .09, .00, and 1.00 respectively. Under quadratic-weighted $\kappa$, the full configuration scores 0.857/1.000/0.500 across Sat-A/Sat-B/Step-drop, compared with 0.250/0.543/0.941 for Action-only; the same regime contrast remains visible.

\clearpage

\section{Prompt Templates and Conservative Execution Policy}
\label{app:artifact}
\label{app:api_access}
\label{app:prompts}

Five of 522 Stage~1 calls remained empty after retry and are treated as parse failures; Appendix~\ref{app:exp1_full} reports the corresponding per-corpus coverage. Stage~1 uses GPT-4o, Claude Sonnet~4.5, and Qwen-Max at temperature 0 with one segment per call and JSON output; few-shot runs use fixed stratified exemplars ($k{=}6$ for FreeNotes and $k{=}5$ for WLP/ONS). Exp~3 uses Claude Sonnet~4.5 with per-file \texttt{SignalFindings} and five independent calls per condition--file pair.

The boxes below reproduce the prompt templates documented in the camera-ready manuscript. The report changes their visual framing, not their text. Placeholders such as \field{\{raw\_text\}} and \field{\{findings\_json\}} stand for runtime inputs; these templates are not complete per-example messages. Where the documented prompt refers to bilingual equivalents in the released code, the original UTF-8 strings must be obtained from those research materials.

\subsection{Stage~1: EDE Prompts}
Stage~1 uses one shared core system message plus one corpus-specific framing paragraph. Zero-shot and few-shot conditions share the same core instruction; the few-shot message prepends the fixed stratified exemplars described after the boxes.

\begin{exactprompt}{Stage 1 - shared core system message}
You are an annotation assistant for the
Epistemic Directive Extraction (EDE) task. Given
a single segment of scientific text, predict
three labels.

# Task overview

For each segment, decide:

1. has_directive (binary, 0 or 1)
   - 1 if the segment carries an actionable
     directive that should influence downstream
     pipeline behavior (changes a parameter,
     flags data quality, suggests an analysis,
     modifies the protocol, or changes a
     condition).
   - 0 if the segment is purely descriptive, a
     passing observation, an introduction, or a
     generic procedural step with no
     decision-changing content.

2. directive_type (5 classes, only when
   has_directive=1; otherwise null)
   - FLAG_DATA: a warning about data validity,
     contamination, exclusion, or quality
     concerns affecting downstream
     interpretation.
   - CONDITION_CHANGE: changing experimental
     conditions, sample types, reagent versions,
     or environmental setup.
   - ANALYSIS_SUGGESTION: a recommendation about
     how to analyze, measure, or examine
     data/samples.
   - PROTOCOL_CHANGE: modifying a procedural step
     (skip, repeat, add, reorder).
   - PARAMETER_SHIFT: changing a numerical
     parameter (time, temperature, volume,
     concentration, voltage).

3. epistemic_status (3 classes, only when
   has_directive=1; otherwise null)
   - FACT: the writer states the directive as a
     definite outcome. No hedging.
   - JUDGMENT: the writer expresses uncertainty,
     qualitative assessment, or a tentative
     interpretation ("seems", "may", "looks
     like"; bilingual hedge markers also listed
     for FreeNotes; see released code for exact
     UTF-8 strings).
   - SUGGESTION: the writer offers an optional or
     recommended action with user discretion
     ("recommend", "optional", "if desired",
     "should consider"; bilingual equivalents in
     released code).

# Output format

Respond with ONLY a single JSON object on one
line, no preamble, no markdown, no explanation:

{"has_directive": 0, "directive_type": null,
 "epistemic_status": null}

or

{"has_directive": 1, "directive_type": "FLAG_DATA",
 "epistemic_status": "JUDGMENT"}

# Constraint

If has_directive=0, both directive_type and
epistemic_status MUST be null. If has_directive=1,
both directive_type and epistemic_status MUST be
non-null.
\end{exactprompt}

\begin{exactprompt}{Stage 1 - FreeNotes corpus context}
# Corpus context

Segments come from informal bilingual
(Chinese/English) nanopore electrophysiology lab
notebooks. Text is short, often colloquial, may
mix Chinese and English in one segment. Authors
frequently use bilingual hedge phrases (English:
"looks like", "unclear", "don't know", "may",
"seem"; Chinese equivalents in released code).
\end{exactprompt}

\begin{exactprompt}{Stage 1 - ONS corpus context}
# Corpus context

Segments come from semi-formal Open Notebook
Science (ONS) entries written in English. Style
is reflective and narrative, with the author
describing their day's work or plans. Many
segments are introductory or summary sentences
without actionable directives.
\end{exactprompt}

\begin{exactprompt}{Stage 1 - WLP corpus context}
# Corpus context

Segments come from published Wet Lab Protocols
(WLP) on protocols.io. Text is formal English in
imperative style. Common markers: "Note:",
"Tip:", "Optional:", "OPTIONAL:", "Safe Stopping
Point:", "DO NOT" (all-caps), "may be kept for
[duration]" (storage spec), "should be X"
(procedure or parameter).
\end{exactprompt}

\begin{exactprompt}{Stage 1 - zero-shot user message}
Segment:
"""
{raw_text}
"""

Predict the JSON tuple.
\end{exactprompt}

\begin{exactprompt}{Stage 1 - few-shot user message}
Here are {k} examples, then your target segment.

# Example 1
Segment:
"""
{exemplar_text_1}
"""
Answer: {"has_directive": 1,
         "directive_type": "FLAG_DATA",
         "epistemic_status": "JUDGMENT"}

# Example 2
...

# Target
Segment:
"""
{raw_text}
"""
Answer:
\end{exactprompt}

Exemplars are sampled once per corpus from the document-stratified train fold (seed 42), with one exemplar per \texttt{directive\_type} class observed in the train fold plus one \texttt{has\_directive=0} negative. The same exemplar set is reused for every test segment. \texttt{CONDITION\_CHANGE} appears only in the FreeNotes train fold, which is why $k{=}6$ for FreeNotes and $k{=}5$ for WLP and ONS.

\subsection{Exp~3: Layer~2 Agent Prompts}
The External-LLM baseline uses the standalone system and user prompt shown below. For the remaining six original conditions, the system prompt is assembled as the shared base plus the condition-specific suffix and loaded representation; executor-enabled conditions additionally append the conservative policy block in Appendix~\ref{app:prompts:rules}. The post-hoc Label-policy baseline makes no additional Layer~2 calls and therefore has no separate prompt.

\begin{exactprompt}{Exp 3 - External-LLM baseline system prompt}
You are an assistant helping a scientist decide
how to process nanopore single-molecule sensing
recording files.

For each file, choose exactly one of these five
actions:
- KEEP_FULL: keep the full recording for
  downstream analysis as-is.
- FLAG_FOR_REVIEW: keep the file but mark it for
  human review before analysis.
- RAISE_THRESHOLD: keep the file but raise the
  event-detection threshold.
- TRUNCATE_AT: truncate the recording at a
  specific timestamp.
- SKIP_FILE: exclude the file from downstream
  analysis.

The notebook below was written by the scientist
during the experiment. It may contain firm
observations, tentative interpretations the
scientist was unsure about, and forward-looking
suggestions for follow-up. Please consider the
author's level of certainty when choosing an
action: act on firm observations, treat tentative
interpretations conservatively, and treat
forward-looking suggestions as advisory only.

Emit your decision by calling the `emit_decision`
tool. Provide a brief rationale.
\end{exactprompt}

\begin{exactprompt}{Exp 3 - External-LLM baseline user message}
Notebook for this experimental session:

{RAW_NOTEBOOK_TEXT}

--

Signal findings for the current file
(`{FILE_ID}`):

{SIGNAL_FINDINGS_JSON}

--

Based on the notebook and the signal findings
above, choose one action for this file by
calling `emit_decision`.
\end{exactprompt}

\begin{exactprompt}{Exp 3 - shared base system prompt}
You are a nanopore signal-processing decision
agent.

Your task: given signal findings extracted from
one ABF (Axon Binary Format) current-trace file,
decide the appropriate ProcessingDecision for
that file by calling the `emit_decision` tool
exactly once.

# Available actions
- KEEP_FULL: use the entire trace as-is.
- TRUNCATE_AT: trim the trace at a timestamp;
  provide `truncate_at_s`.
- RAISE_THRESHOLD: use elevated event-detection
  threshold; provide `threshold_multiplier`.
- FLAG_FOR_REVIEW: include in analysis but flag
  the file for human review.
- SKIP_FILE: exclude the file entirely from
  analysis.

The five actions form an ordered severity scale:
KEEP_FULL < FLAG_FOR_REVIEW < RAISE_THRESHOLD
< TRUNCATE_AT < SKIP_FILE.
\end{exactprompt}

\begin{exactprompt}{Exp 3 - Action-only suffix}
# Reasoning protocol -- flat directive list

A flat list of author directives is loaded below.
There are NO decision_points, NO
recovery_strategies, NO parameter_space, NO
trigger_conditions. Each directive has:
  - `action` and `action_parameters`
  - `epistemic_status` in {FACT, JUDGMENT,
    SUGGESTION}
  - `confidence_weight` in {1.0, 0.6, 0.3}
  - `uncertainty_markers` (verbatim hedges)
  - `raw_excerpt` (verbatim source text)
  - `source_segment` (provenance label)

Examine each directive against signal_findings to
determine whether it applies to the current file.
Apply matching directives weighted by their
confidence. The author's epistemic confidence
should inform how decisively you act.

# Executable action mapping
  - `truncate_at_step_drop`     -> TRUNCATE_AT
  - `flag_full_file_uncertain`  -> FLAG_FOR_REVIEW
  - `flag_remainder_uncertain`  -> RAISE_THRESHOLD
  - `raise_threshold_in_window` -> RAISE_THRESHOLD
  - `confirm_events_observable` -> ADVISORY
  - `cross_read_comparison`     -> ADVISORY
  - `split_processing_phases`   -> ADVISORY

When multiple directives apply, the most severe
action wins.

# Directives (loaded for this run)
\end{exactprompt}

\begin{exactprompt}{Exp 3 - Raw-notes suffix}
# Reasoning protocol -- notebook excerpts

A Markdown document containing curated notebook
excerpts is included below. Each excerpt has a
`source_segment` label and the original text. You
have no other information about the author's
intent.

Infer cautiously by reading the excerpts together
with the numerical signal findings. Do not assume
that any excerpt is an executable instruction;
treat each as a piece of evidence from the lab
record. Decide which excerpts (if any) apply to
the current file by matching language cues
(voltage values, time references, descriptive
judgments) against signal_findings.

When you call `emit_decision`, list the relevant
`source_segment` labels (e.g., `SessionA_s020`) in
`contributing_directives`. If no excerpt clearly
applies, leave `contributing_directives` empty.

# Excerpts (loaded for this run)
\end{exactprompt}

\begin{exactprompt}{Exp 3 - Notes2Skills skill suffix}
# Reasoning protocol -- Markdown skill

A Markdown skill document is included below. Read
it as natural-language guidance for handling the
current `.abf` file. The document defines:
  - the action vocabulary used in this domain;
  - the core decision principle;
  - per-directive guidance with the author's
    verbatim notebook excerpts, an epistemic
    interpretation, a `default_action`, and
    optionally `candidate_action`(s) that require
    human review;
  - a decision policy and a list of known failure
    modes.

# Tool action mapping

The `emit_decision` tool accepts these five
actions: KEEP_FULL, TRUNCATE_AT, RAISE_THRESHOLD,
FLAG_FOR_REVIEW, SKIP_FILE.

The Markdown skill also uses `KEEP_FULL_WITH_NOTE`
as the default action for context directives.
When the appropriate decision is
`KEEP_FULL_WITH_NOTE`, emit `KEEP_FULL` and place
the contextual note content in the `rationale`
field of `emit_decision`.

# Directive identifiers

Each directive has a `display_id` of the form
`D01..Dn` (file-level) or `C01..Cm` (context).
Use these IDs in the `contributing_directives`
field of `emit_decision`.

# Skill (loaded for this run)
\end{exactprompt}

\begin{exactprompt}{Exp 3 - common user message}
signal_findings for the current ABF file:

```json
{findings_json}
```

Please reason briefly about which directive(s)
(if any) apply to this file and what the
appropriate ProcessingDecision is. Then call
`emit_decision` exactly once with your final
answer.
\end{exactprompt}

\subsection{Conservative Execution Rules}
\label{app:prompts:rules}
For the three executor-enabled conditions (Action-only + executor, \systemname{} + verify, and \systemname{} + verify+elevate), the following policy text is appended to the Layer~2 system prompt as the final block. The wording below is the policy content used by those conditions; only typographic formatting has been added.

\begin{exactpolicy}{Executor-enabled conservative policy block}
\textbf{Purpose.} These rules govern how the Layer~2 agent translates notebook-derived skill content into a single downstream processing action. They are designed to prevent the conversion of epistemic uncertainty into automated intervention while preserving the agent's ability to follow explicit operational directives. They are not designed to maximize ordinary classification accuracy on any particular session.

\medskip\noindent\textbf{Action vocabulary.} The agent must select exactly one action from \{\texttt{KEEP\_FULL}, \texttt{FLAG\_FOR\_REVIEW}, \texttt{RAISE\_THRESHOLD}, \texttt{TRUNCATE\_AT}, \texttt{SKIP\_FILE}\}. When \texttt{TRUNCATE\_AT} is selected, the agent must report the most specific time point or event boundary supported by the skill content or signal findings; it must not invent a boundary not present in the input.

\medskip\noindent\textbf{Core principle.} The agent distinguishes two categories of skill content:
\begin{itemize}[leftmargin=*,nosep]
  \item \emph{Epistemic uncertainty}: the skill or notebook indicates that the evidence is weak, tentative, ambiguous, low-confidence, unresolved, questionable, or otherwise uncertain.
  \item \emph{Explicit operational directive}: the skill or notebook provides a concrete processing instruction, identified by at least one of a specific time point or event boundary, a specific threshold value or adjustment magnitude, a clearly localized window for intervention, or an explicit non-target file designation.
\end{itemize}
\textbf{Critical rule}: epistemic uncertainty alone does not authorize \texttt{TRUNCATE\_AT}, \texttt{RAISE\_THRESHOLD}, or \texttt{SKIP\_FILE}. These actions require an explicit operational directive. When the skill expresses uncertainty without an accompanying operational directive, the agent must select \texttt{FLAG\_FOR\_REVIEW}.

\medskip\noindent\textbf{Binding rules.}
\begin{description}[leftmargin=*,style=nextline]
  \item[Rule 1 -- \texttt{SKIP\_FILE}.] Select \texttt{SKIP\_FILE} only when the file is identified as outside the scope of downstream experimental processing: a reference, control, calibration, or blank recording; a procedural or auxiliary role (system preparation, cleaning, flushing); explicitly marked as not part of the experimental target set; or described as serving a different experimental role rather than as an uncertain experimental result. Do \emph{not} select \texttt{SKIP\_FILE} when the signal is weak or noisy, the evidence is uncertain, the notebook expresses hesitation or recommends human inspection, or the signal quality is imperfect but the file remains potentially informative. Route such cases to \texttt{FLAG\_FOR\_REVIEW}.

  \item[Rule 2 -- \texttt{TRUNCATE\_AT}.] Select \texttt{TRUNCATE\_AT} only when the skill provides an explicit truncation basis, requiring all of: (i) a specific time point, event boundary, or before/after transition; (ii) a concrete reason localized to the post-boundary region (contamination, saturation, artifact, post-event failure); (iii) the skill identifies truncation as a needed processing step, not a tentative possibility. Do \emph{not} select \texttt{TRUNCATE\_AT} when later signal is described as uncertain without an identified boundary, the notebook describes general degradation without specifying when it begins, or hedged language is present without an accompanying explicit boundary and concrete operational reason.

  \item[Rule 3 -- \texttt{RAISE\_THRESHOLD}.] Select \texttt{RAISE\_THRESHOLD} only when the skill provides at least one of a specific threshold value, a specific adjustment magnitude, a clearly localized window for stricter thresholding, or a direct operational instruction to filter weak or noisy events. Conditional formulations such as ``consider raising threshold'' or ``may need a stricter cutoff'' without a specific value, magnitude, or localized window route to \texttt{FLAG\_FOR\_REVIEW}.

  \item[Rule 4 -- \texttt{FLAG\_FOR\_REVIEW}.] The default conservative action when the skill expresses epistemic uncertainty and none of Rules 1--3 are satisfied by an explicit operational directive. Valid reasons include weak evidence, tentative interpretation, ambiguous signal quality, low-confidence event identification, unresolved disagreement between signal evidence and notebook interpretation, insufficient evidence for automatic exclusion, truncation, or threshold adjustment, a notebook recommendation that human inspection is needed, or hedged language without a concrete directive. Selecting \texttt{FLAG\_FOR\_REVIEW} preserves the file for human or downstream expert inspection.

  \item[Rule 5 -- \texttt{KEEP\_FULL}.] Select \texttt{KEEP\_FULL} only when the file appears usable as a complete recording, no unresolved uncertainty is expressed in the skill for this file, no explicit intervention is specified by Rules~1--3, and the file is not identified as a non-target recording.
\end{description}

\medskip\noindent\textbf{Hedged language.} The terms below indicate epistemic uncertainty. The list is illustrative; the agent treats any natural-language expression of epistemic uncertainty as covered, not only the listed terms.

\begin{center}\scriptsize
\setlength{\tabcolsep}{2pt}
\renewcommand{\arraystretch}{1.05}
\begin{tabularx}{0.98\columnwidth}{@{}XXX@{}}
\toprule
\textbf{Modal/conditional} & \textbf{Evaluative} & \textbf{Limiting} \\
\midrule
may, might, could & weak, tentative, questionable & not sure, unclear \\
consider, possibly, maybe & ambiguous, low confidence & needs checking, need to inspect \\
if needed, if appropriate & seems, appears & uncertain, unresolved \\
\bottomrule
\end{tabularx}
\end{center}

When such language is present without an accompanying explicit operational directive (a specific timestamp, threshold value, adjustment magnitude, localized window, or non-target file designation), the agent must not select \texttt{TRUNCATE\_AT}, \texttt{RAISE\_THRESHOLD}, or \texttt{SKIP\_FILE}; the correct selection is \texttt{FLAG\_FOR\_REVIEW}.

\medskip\noindent\textbf{Prohibited decision patterns.}
\begin{enumerate}[leftmargin=*,nosep]
  \item \emph{Uncertainty-to-intervention escalation}: selecting \texttt{TRUNCATE\_AT}, \texttt{RAISE\_THRESHOLD}, or \texttt{SKIP\_FILE} solely because evidence is weak or uncertain.
  \item \emph{Hedge-as-command interpretation}: treating phrases such as ``might raise threshold'' or ``consider truncating'' as mandatory operational instructions.
  \item \emph{Review-to-discard conversion}: treating ``needs review'' or ``requires inspection'' as a reason to skip, truncate, or raise threshold.
  \item \emph{Boundary invention}: selecting \texttt{TRUNCATE\_AT} without a specific time point or event boundary present in the skill or findings.
  \item \emph{Threshold invention}: selecting \texttt{RAISE\_THRESHOLD} without a concrete threshold value, adjustment magnitude, localized window, or explicit thresholding directive present in the skill or findings.
  \item \emph{Always-\texttt{FLAG} collapse}: selecting \texttt{FLAG\_FOR\_REVIEW} for every file, including files for which an explicit \texttt{SKIP\_FILE}, \texttt{TRUNCATE\_AT}, or \texttt{RAISE\_THRESHOLD} directive is present in the skill content.
\end{enumerate}

\medskip\noindent\textbf{Scope.} These rules govern action selection for files whose skill content can be evaluated against a single, internally consistent epistemic state. The document does not specify behavior for files whose skill content is internally contradictory between two operational directives, for files for which signal findings and skill directives point to incompatible actions, or for files for which the skill content is empty or absent. In such cases the agent defaults to \texttt{FLAG\_FOR\_REVIEW}; the basis is unresolved skill content rather than evaluated epistemic uncertainty.
\end{exactpolicy}

\end{document}